\AtBeginDocument{%
  \paperwidth=\dimexpr
    1in + \oddsidemargin
    + \textwidth
    + 1in + \oddsidemargin
  \relax
  \paperheight=\dimexpr
    1in + \topmargin
    + \headheight + \headsep
    + \textheight
    + 1in + \topmargin
  \relax
  \usepackage[pass]{geometry}\relax
}

\documentclass[smallcondensed]{svjour3}     
\smartqed  

\usepackage{array,multirow,graphicx}
 \usepackage{float}
\usepackage{longtable}
\usepackage{csquotes}
\usepackage{booktabs}
\usepackage{url} 
\usepackage{subcaption}
\usepackage{latexsym}
\usepackage{amssymb}
\usepackage{mdwlist}
\usepackage [misc]{ifsym}
\usepackage{nccmath}
\usepackage{pdflscape}
\usepackage[numbers]{natbib}
\usepackage[bottom]{footmisc}
\usepackage{xcolor}
\usepackage{color, soul}
\usepackage{amsthm}
\theoremstyle{definition}
\usepackage{ulem}
\usepackage{comment}
\usepackage{pifont}
\usepackage{bbding}
\usepackage{afterpage}
\usepackage{ragged2e}
\usepackage{array}

\usepackage{color}
\usepackage{soul}

\newcommand*\colourcheck[1]{%
  \expandafter\newcommand\csname #1check\endcsname{\textcolor{#1}{\ding{52}
  }}%
}
\colourcheck{black}
\colourcheck{gray}

\usepackage{enumitem}
\newcommand{\tablistcommand}{
            \leavevmode\par\vspace{-\baselineskip}
                            }
\newlist{tableitems}{itemize}{1}
\setlist[tableitems]{nosep,     
                     topsep     = 0pt               ,
                     partopsep  = 0pt               ,
                     leftmargin = *                 ,
                     label      = \textbf{-}        ,
                     before     = \tablistcommand   ,
                     after      = \tablistcommand
                     }

\title{comparison table}
\begin{document}
\title{Conversational Question Answering: A Survey}

\author{
\thanks{*The corresponding authors.}
{Munazza Zaib*\thanks{\Letter { munazza-zaib@hdr.mq.edu.au}}} 
\and Wei Emma Zhang \and Quan Z. Sheng* \thanks{\Letter{ michael.sheng@mq.edu.au}}  \and Adnan Mahmood \and Yang Zhang
}

\authorrunning{Munazza Zaib et al.} 

\institute{Munazza Zaib, Quan Z. Sheng, Adnan Mahmood, Yang Zhang \at
              Department of Computing, Faculty of Science and Engineering, Macquarie University, NSW 2109, Australia \\
                        \and
           Wei Emma Zhang  \at
              School of Computer Science, The University of Adelaide, North Terrace, Adelaide, SA 5005, Australia
}

\date{Received: date / Accepted: date}

\maketitle

\begin{abstract}
Question answering (QA) systems provide a way of querying the information available in various formats including, but not limited to, unstructured and structured data in natural languages. It constitutes a considerable part of conversational artificial intelligence (AI) which has led to the introduction of a special research topic on \textit{Conversational Question Answering} (CQA), wherein a system is required to understand the given context and then engages in multi-turn QA to satisfy a user’s information needs. Whilst the focus of most of the existing research work is subjected to single-turn QA, the field of multi-turn QA has recently grasped attention and prominence owing to the availability of large-scale, multi-turn QA datasets and the development of pre-trained language models. With a good amount of models and research papers adding to the literature every year recently, there is a dire need of arranging and presenting the related work in a unified manner to streamline future research. This survey, therefore, is an effort to present a 
comprehensive review of the state-of-the-art research trends of CQA primarily based on reviewed papers from 
2016-2021. Our findings show that there has been a trend shift from single-turn to multi-turn QA which empowers the field of Conversational AI from different perspectives. This survey is intended to provide an epitome for the  research 
community with the hope of laying a strong foundation for the field of CQA.


\end{abstract}

\section{Introduction}
\label{intro}
Designing an intelligent dialog system that not only matches or surpasses a human’s level on carrying out an interactive conversation, but also answers the questions on a variety of topics, i.e., ranging from recent news about NASA to a biography of a famous political leader, has been one of the outstanding goals in the field of artificial intelligence (AI) \cite{DBLP:journals/ftir/GaoGL19}. 
A 
quickly-increasing 
number of research papers prove the promising potential and the growing interest of researchers from both academia and industry in conversational AI.

Conversational AI constitutes an integral part of Natural User Interfaces \cite{DBLP:journals/ftir/GaoGL19} and is attracting 
significant attention from researchers in Information Retrieval (IR), Natural Language Processing (NLP), and Deep Learning (DL) communities. For example, 
AAAI 2020 introduced a
 special workshop focusing on ``Reasoning for Complex Question Answering’', that featured a special focus on machine intelligence and common sense reasoning. Similarly, SIGIR 2018 introduced a new track entitled ``Artificial Intelligence, Semantics and Dialog’' to bridge the gap between IR and AI. The track is especially focused on QA, Conversational Dialog Agents, and Deep Learning for IR and agents. One of the top conferences in NLP, EMNLP, has had a track called ``Information Retrieval and Question Answering'' for years and from 2019 
 it has started inviting papers for the field of ``Question Answering'' as a separate track owing to the increasing research interests of the community and its faced-paced growth.

\begin{figure}[htb!]
\center
  \includegraphics[width=0.745\textwidth]{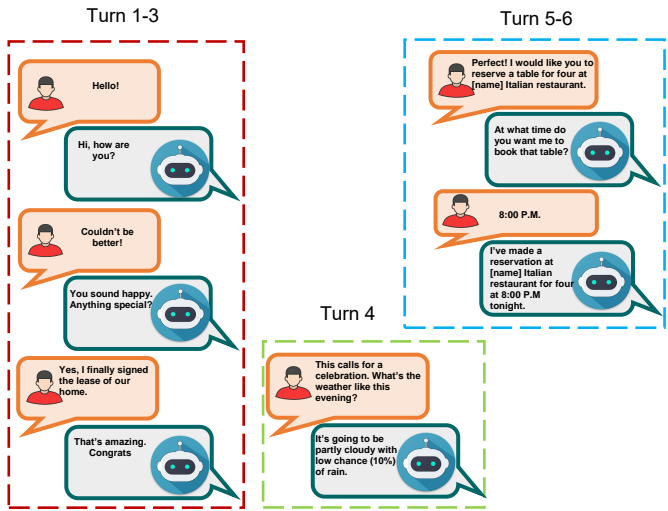}
  \caption{Categorizations of conversational AI. Turn 1-3 depict chat-oriented dialog system, turn 4 portrays the element of QA dialog system, and turn 5-7 reflects the task-oriented conversation.}
  \label{fig:types}
\end{figure}

The 
field of conversational AI can be segregated into three groups namely, i) \textit{task-oriented dialog systems} that are required to perform tasks on the users' behalf such as making a reservation in a restaurant or scheduling an event, ii) \textit{chat-oriented dialog systems} that need to carry out a natural and interactive conversation with the users, and iii) \textit{QA dialog systems} that are responsible to provide 
clear and concise answers to the users' questions based on information deduced from different data sources such as text documents or knowledge bases. The examples of each of the aforementioned categories are given in Fig.~\ref{fig:types}. The conversation shown in Fig.~\ref{fig:types} comprises of multiple turns and each turn consists of a question and an answer \cite{DBLP:journals/tacl/ReddyCM19}.


The chat-oriented and task-oriented dialog systems have been well-researched topics resulting in a number of successful
dialog agents such as Amazon Alexa\footnote{https://www.amazon.com.au/b?node=5425666051}, Apple Siri\footnote{https://www.apple.com/au/siri/}, and Microsoft Cortana\footnote{https://www.microsoft.com/en-us/cortana}. However, QA dialog systems are fairly new and still require extensive research. 
Many QA challenges 
have been identified and initial solutions have been proposed \cite{DBLP:conf/emnlp/RajpurkarZLL16,DBLP:journals/tacl/KociskySBDHMG18,DBLP:conf/acl/JoshiCWZ17,DBLP:conf/naacl/SusterD18,DBLP:conf/emnlp/ZellersBSC18,DBLP:journals/pvldb/CuiXWSHW17,DBLP:conf/coling/BaoDYZZ16,DBLP:conf/semweb/TrivediMDL17}, 
giving the rise 
of \textit{Conversational Question Answering} (CQA). 
CQA techniques form the building blocks of QA dialog systems. The idea behind CQA is to ask the machine to answer a question based on the provided passage and this, in turn, has the potential to revolutionize the way 
humans interact with the machines. However, this interaction could turn into a multi-turn conversation if a user requires more detailed information about the question. 
%
The notion of CQA can be thought of as a simplified but concrete conversational search setting \cite{DBLP:conf/sigir/Qu0QCZI19}, wherein the system returns one correct answer to a user's question instead of a list of relevant documents or links as is the case with traditional search engines. The top search engine companies such as Microsoft and Google have incorporated CQA into their mobile-based search engines (also known as \textit{digital assistants}) to improve the users' experience when interacting with them.

CQA is an effective way for humans to gather information and is considered as a benchmark task to evaluate a machine's capability to understand and comprehend the input provided in written natural language \cite{DBLP:journals/corr/abs-1812-03593}. Such CQA systems have significant applications in areas like customer service support \cite{DBLP:conf/acl/CuiHWTDZ17} or QA dialog systems \cite{DBLP:journals/tacl/ReddyCM19, DBLP:conf/emnlp/HewlettJLG17}. The task of CQA poses several challenges to the researchers hence resulting in considerable interesting yet innovative researches over the past few years.  

\vspace{2mm}
\noindent\textbf{Papers' Selection:} The research papers reviewed in this survey are high quality papers selected from the top
NLP and AI conferences, including but not limited to, ACL\footnote{https://www.aclweb.org/}, SIGIR\footnote{https://sigir.org/}, NeurIPS\footnote{https://nips.cc/}, NAACL\footnote{https://naacl.org/}, EMNLP\footnote{https://sigdat.org/}, ICLR\footnote{https://iclr.cc/}, AAAI\footnote{https://www.aaai.org/}, IJCAI\footnote{https://www.ijcai.org/}, CIKM\footnote{http://www.cikmconference.org/}, SIGKDD\footnote{https://www.kdd.org/}, and WSDM\footnote{http://www.wsdm-conference.org/}. 
Other than 
published research papers in the aforementioned conferences, we have also considered good papers in e-Print archive\footnote{https://arxiv.org/} as they manifest the latest research outputs. We selected papers from archive 
using 
three metrics:
paper quality, method novelty, and the number of citations (optional).

\begin{figure}[!tb]
\begin{subfigure}{.5\textwidth}
  \centering
  \includegraphics[width=.95\linewidth]{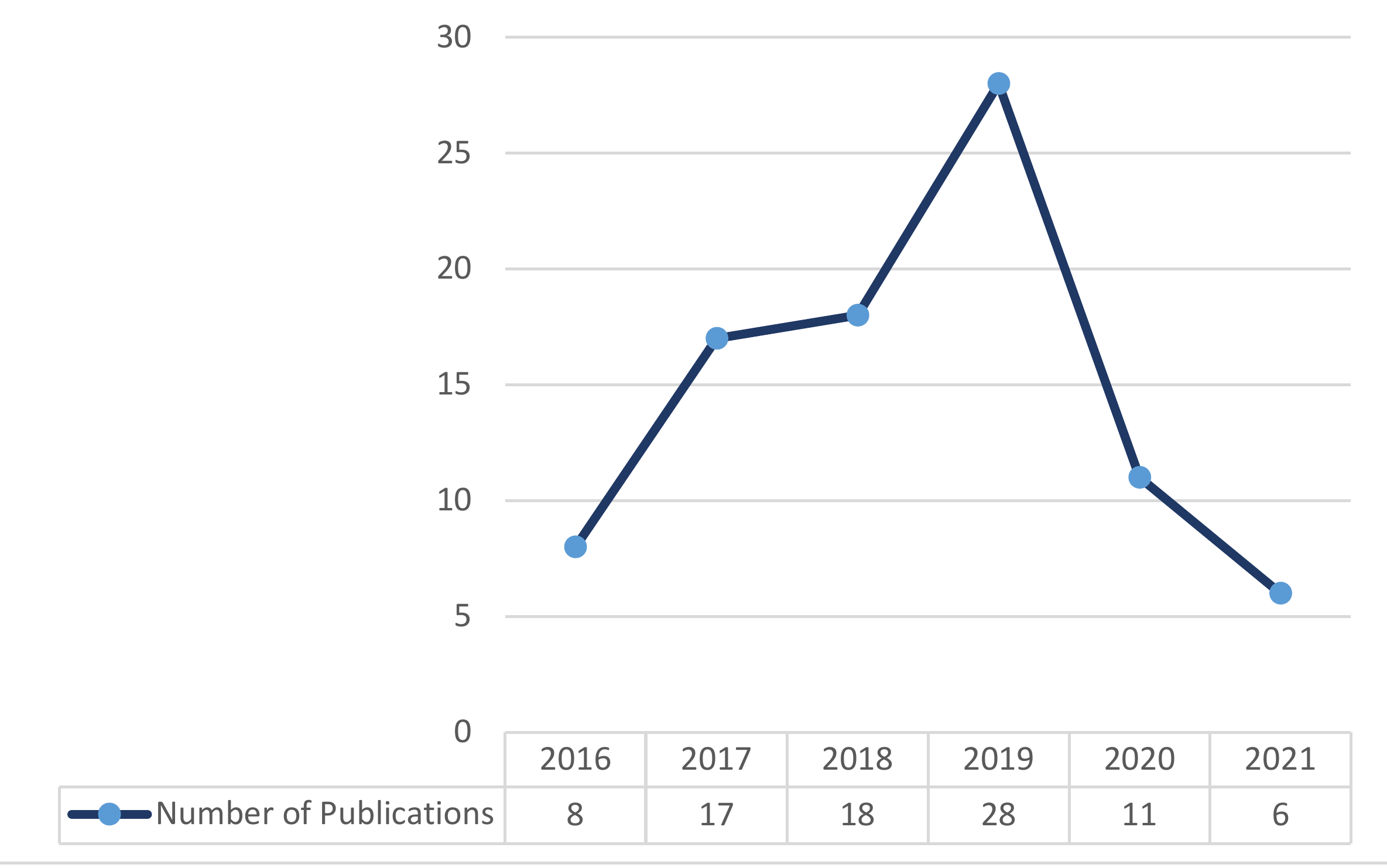}
  \caption{Year-wise distribution.}
  \label{fig:yearwise}
\end{subfigure}%
\begin{subfigure}{.5\textwidth}
  \centering
  \includegraphics[width=.95\linewidth]{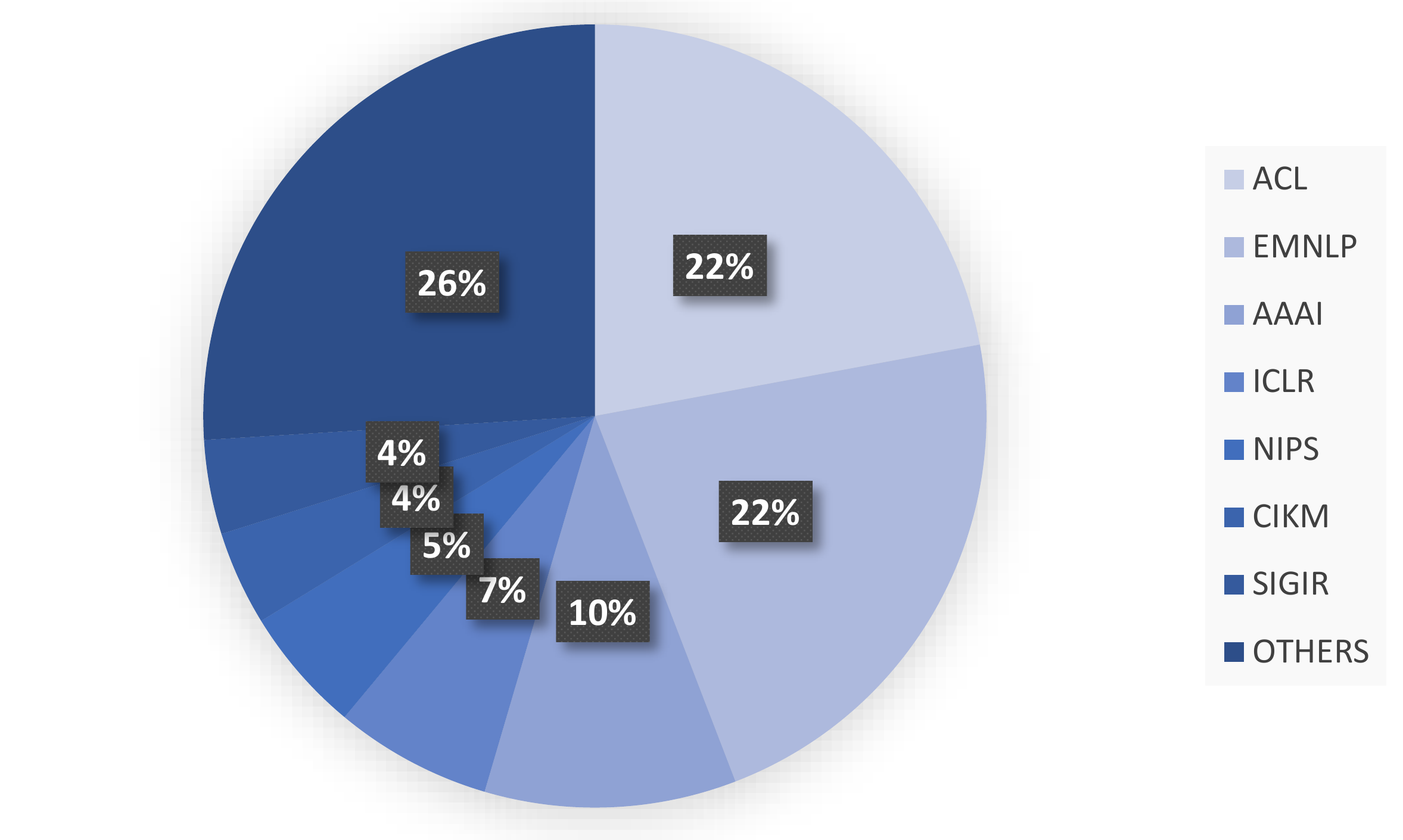}
  \caption{Venue distribution.}
  \label{fig:con}
\end{subfigure}
\caption{(a) Year-wise statistics of the selected survey papers between 2016 and 2021 inclusive. The figure depicts that the field of CQA saw its rise recently. (b) Venue-wise distribution of the reviewed research works.}
\label{fig:conf}
\end{figure}

Fig.~\ref{fig:yearwise} depicts the year-wise distribution of the papers 
reviewed in our survey. Our survey encompasses over 
80 
top-notch conferences and journal papers. The number of papers pertinent to CQA steadily increase from year 2016 onwards, with the highest being in 2019. Coincidentally, 2019 also marks the year when the fields of natural language generation and natural language understanding were revolutionized with the introduction of pre-trained language models. These pre-trained language models have the potential to address the issue of data scarcity and bring considerable advantages
by generating contextualized word embeddings \cite{45663f8dbad442a7b649001bfeb2be72}. 
This rise of interest depicts the gradual shift in focus of the researchers in both academia and industry in utilizing pre-trained language models for the design of CQA systems. Also, Fig.~\ref{fig:con} portrays the venue-wise distribution of the research works we have reviewed, with ACL and EMNLP being the top 
venues for natural-language related progress. We note, though, that more than 25\% of papers come from  a variety of conferences/journals outside of the typical venues further attesting to the fact that this is an interdisciplinary topic spanning different areas such as 
knowledge management, knowledge discovery, information retrieval, and artificial intelligence.

\vspace{2mm}
\noindent{\bf What Makes This Survey Different?}
There have been several published literature reviews on QA systems, i.e., in the context of both machine reading comprehension and KB-QA. In \cite{DBLP:journals/ftir/GaoGL19}, the authors provide 
an overview of Conversational AI with a detailed discussion of neural methods and deep learning techniques being used in designing efficient conversational agents. These conversational agents include task-oriented dialog systems, chat-oriented dialog systems, and QA dialog systems. Although the paper sheds some light on several research works and datasets pertinent to CQA, 
it does not cover the recent trends and methods on CQA.
The authors in \cite{fu2020survey} recently published their literature review which primarily highlights the complex QA over knowledge bases. The paper covers all the datasets and different approaches that are employed in complex KB-QA systems along with the discussion of complex QA. CQA is just mentioned as a ``future trend'' with 
minimal discussion.
A summary of the techniques and methods of single-turn QA is presented in \cite{liu2019neural} along with proposing a general modular architecture needed for it. The paper further discusses the techniques that could be used in each module. Again, CQA is discussed briefly as a newly emerging trend along with the different challenges.
Another recent effort, \cite{gupta-etal-2020-conversational}, delineates on the latest trends and methods to cater to the successful implementation of multi-turn machine reading comprehension. However, it lacks the discussion on other forms of multi-turn QA.

The key aspect that makes this survey to stand out among its predecessors is its focus on CQA encompassing both sequential KB-QA and conversational machine reading comprehension. 
Multi-turn QA has been discussed very nominally in previous surveys. It is an essential aspect to consider when discussing the process of carrying out a natural conversation with a machine. Based on the review, we thoroughly discuss the research works 
of CQA, the techniques employed, and highlight the merits and demerits of 
different techniques. Finally, we highlight and discuss existing challenges related to the field of CQA along with an attempt to suggest some application areas.

The field of CQA is witnessing its golden era in terms of research publications and this calls for having a strong background work that discusses its challenges and trends as a separate field than single-turn QA. Thus, this survey is an effort to establish a strong foundation for CQA which would benefit 
the research communities as well. This work provides detailed insights into important ideas pertinent to CQA systems that are needed to design interactive and engaging conversational systems. 
To the best of our knowledge, this is the first work to 
investigate the field of CQA in detail. 
We hope that this paper would turn out to be a valuable resource for researchers who are interested in this area.  

\vspace{2mm}
\noindent{\bf The Survey Structure}.
The rest of the paper is organized as follows. 
 %
 Section~\ref{section2} delineates on a brief background of single-turn QA and leads the discussion on CQA. This section further highlights the categorization of CQA systems based on the source they utilize to answer the questions.
Section~\ref{seq kb-qa} 
describes the task of sequential knowledge-based question answering (KB-QA)  system and the general architecture it employs. The section further highlights the techniques used in each module of the system to effectively carry out the task of sequential QA.
Section~\ref{cmrc} describes the task of Conversational Machine Reading Comprehension (CMRC) and how it differs from typical machine reading comprehension. The section also describes how the general architecture of machine reading comprehension can be adapted for conversational machine reading comprehension. It further describes the decomposition of the architecture in different modules and techniques employed in each 
of them.
Section~\ref{dataset} describes the datasets introduced to further improve the work in the field of CQA along with a qualitative comparison of each of them. 
Section~\ref{challenges} highlights the potential applications of the CQA systems in commercial areas along with the research trends that should be explored to leverage the strength of these systems more effectively.
Finally, 
Section~\ref{conclusion} 
offers some concluding remarks.

\section{Conversational Question Answering}
\label{section2}
\begin{figure}[tb!]
\center
\includegraphics[width=\textwidth]{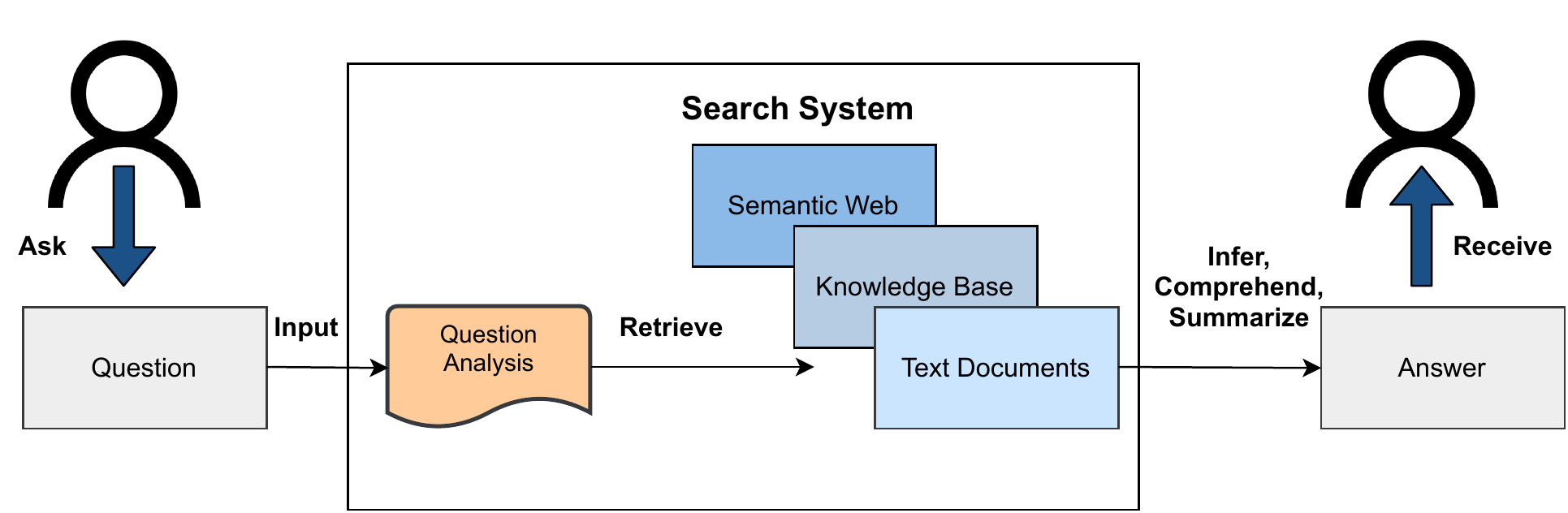}
\caption{The high-level or generic architecture of QA systems where search system corresponds to the different sources. The specific architecture of a QA system depends on the underlying data source.}
\label{highlevel}
\end{figure}

Question answering in general
involves accessing different data sources to find the correct answer for 
an asked question, as depicted in Fig.~\ref{highlevel}. 
It dates back to the 1960s \cite{monz2011machine} when early QA systems, due to rule-based methods and absurdly small size of available datasets, did not achieve well, thereby making it difficult to 
be used 
in practical applications. These systems saw their rise in 2015 and this largely was 
associated with two driving factors:
\begin{itemize}
\item [(a)] The use of deep learning methods to capture the critical information in QA tasks that outperform the traditional rule-based models, and 
\item [(b)] The availability of several large-scale datasets,
i.e., SQuAD \cite{DBLP:conf/emnlp/RajpurkarZLL16}, Freebase \cite{bollacker2008freebase}, MS MARCO \cite{DBLP:conf/nips/NguyenRSGTMD16}, DBpedia \cite{lehmann2015dbpedia}, and CNN \& DAILY MAIL \cite{DBLP:conf/conll/NallapatiZSGX16}, which make it possible to deal with the task of QA on neural architectures more efficiently and further provide a test bed for evaluating the performance of these models.
\end{itemize}

To 
realize the QA tasks more 
close 
to the real-world scenarios, 
several advanced research directions 
have emerged recently. 
One such 
direction is CQA \cite{liu2019neural}, 
which introduces a new dimension of dialog systems that combines the elements of both chit-chat and QA. CQA is a \textit{system ask, user respond} kind of setting where the system can asks a user multiple questions to understand the user's information need \cite{zhang2018towards}. Usually, a user starts the conversation with a particular question in mind and the system searches its database to find an appropriate solution to that query. This could turn into a multi-turn conversation if the user needs to have more detailed information about the topic.
 
\subsection{Categorization of CQA Systems}
\begin{figure}[tb!]
\center
\includegraphics[width=\textwidth]{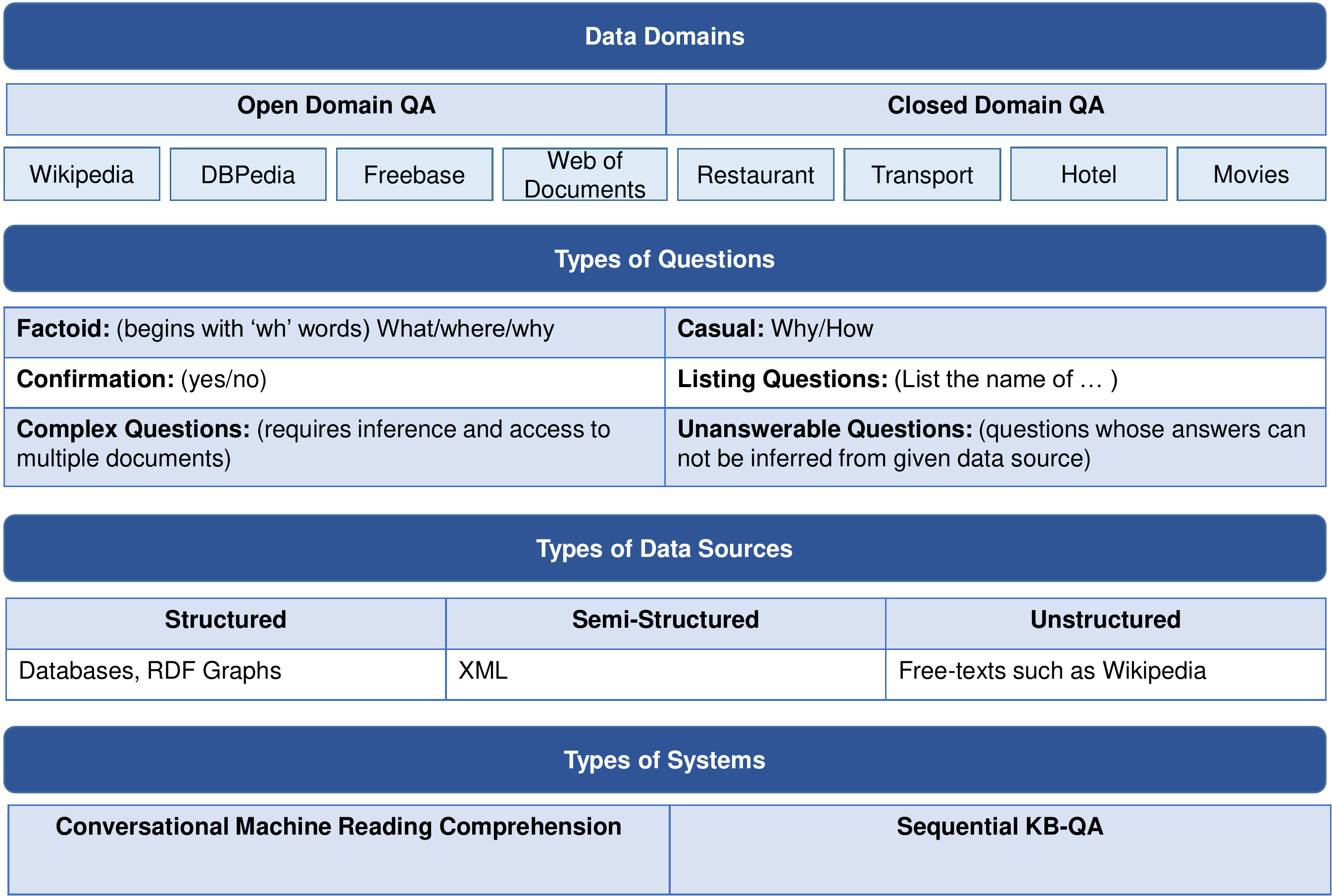}
\caption{Categorization of 
CQA on the basis of: i) data domains, 
ii) types of questions, iii) types of data sources, and iv) types of systems \cite{mishra2016survey}.}
\label{categorization}
\end{figure}

There are several ways of structuring the different aspects of a QA system. Since CQA is categorized as a sub-category of QA, 
the same categorization can be used for CQA systems as well. The categorization of the CQA model could be realized on the basis of the data domain, types of questions, types of data sources, and the types of systems that we are building for the questions at hand \cite{mishra2016survey}. Fig.~\ref{categorization} manifests the possible options that could be utilized to structure 
a CQA 
system. The 
details of each of the category
are given in the rest of this section. 

\subsubsection{Data Domains}

Questions asked by 
users are 
either open domain \cite{jiang-etal-2019-freebaseqa, yang-etal-2015-wikiqa}
in which questions are domain-free and in a broad range, 
or restricted 
to specific 
application domains 
(i.e., closed domain) 
such as Travel \cite{beaver2020towards}, Restaurants \cite{budzianowski2018multiwoz}, Movies \cite{DBLP:journals/corr/abs-1908-03180}, 
and 
Hospitals \cite{budzianowski2018multiwoz}. 
The 
question repository of closed domain 
question answering 
is 
smaller compared 
to open domain 
question answering. 
This makes the models designed for closed domain QA less transferable 
than 
the models for open domain QA. 

It should be noted that the sub-categories of open-domain QA and closed-domain QA are the examples of generic and task-specific datasets.

\subsubsection{Types of Questions}

Questions can be easily classified into various categories primarily depending upon their complexity, the nature of the response, or the techniques that should be utilized to answer them \cite{mishra2016survey}. The classification based on the questions commonly asked by the users is delineated as follows:

\paragraph{Factoid Questions:} Questions which expect the system to find a simple and fact-based answer in a short sentence, e.g., ``\textit{who acted as Chandler in FRIENDS?}''. Factoid questions typically begin with a \textit{wh}-word. Different extraction techniques can be employed to find the answers to the factoid questions. The techniques first recover latent or hidden information in the given question, and then look for the answer in the given text using either structure matching \cite{DBLP:conf/acl/ShenK06} or reasoning \cite{DBLP:conf/emnlp/IyyerBCSD14}. FreebaseQA \cite{jiang-etal-2019-freebaseqa} is one of the examples of factoid QA dataset.

\paragraph{Confirmation Questions:}
Questions which require the answer in a binary format i.e., yes or no, e.g., ``\textit{Is Sydney the capital of Australia?}''. As the answers are not simple extractive text spans from the given source, 
a strong inference mechanism is needed to deduce the answers of confirmation type questions \cite{mishra2016survey}. While there may be a lot of information given about a topic, analyzing if the original statement is true or not is still a challenging task.

\paragraph{Simple Questions:}
Simple questions require small piece of text to find an answer and, thus, they are easier to comprehend. For instance, for a question like ``\textit{What is the magnitude of earthquake in Pakistan?}'', it can easily be deduced that the answer of this question would be a simple numeric value. The process of finding an answer to a simple question consists of three basic steps: i) question analysis, ii) relevant documents/knowledge graphs retrieval, and iii) answer extraction \cite{bouziane2015question}. MS MARCO \cite{DBLP:conf/nips/NguyenRSGTMD16}, SQuAD \cite{DBLP:conf/emnlp/RajpurkarZLL16}, and FreebaseQA \cite{jiang-etal-2019-freebaseqa} are some of the examples of simple questions-based datasets. 

\paragraph{Complex Questions:}
Complex questions are questions that require different types of knowledge or several steps to answer. 
They are difficult to answer and require access to multiple documents or multiple interactions with the system \cite{bhutani-etal-2020-answering}. Complex question like ``\textit{how many cities in China have more population than New Delhi?}'' requires the system to first figure out the population of New Delhi and then compare it with the population of different cities in China. Thus, answering complex questions requires complex techniques such as iterative query generation \cite{DBLP:conf/emnlp/QiLMWM19}, multi-hop reasoning \cite{xiong2021answering}, decomposition into sub-questions \cite{DBLP:conf/acl/IyyerYC17}, and combining cues from the multiple documents \cite{DBLP:conf/sigir/LuPRAWW19}.
LC-QuAD \cite{DBLP:conf/semweb/TrivediMDL17} and CSQA \cite{DBLP:conf/aaai/SahaPKSC18} are some of the examples of complex QA datasets.

\paragraph{Casual Questions:}
Casual questions require detailed explanation pertinent to the entity and they usually start with the words like \textit{why} or \textit{how}.
The answers generated for casual questions are not straight-forward or concise. This generation of detailed answers call for advanced natural language processing techniques that are able to understand the question on different levels of technicality such as semantics and syntax \cite{higashinaka-isozaki-2008-corpus}. An example of such questions could be ``\textit{why do earthquakes occur?}''.

\paragraph{Listing Questions:} These are the 
questions which require the list of entities or facts as an answer, e.g., ``\textit{list the name of all the former presidents of America}''. The techniques that are utilized to answer factoid question works well for the listing questions.
The reason being that QA systems treat such questions as a sequence of factoid questions asked iteratively \cite{mishra2016survey}. 

\paragraph{Unanswerable Questions:} These are the 
questions whose answers cannot be found or deduced via the source text.
Unanswerable questions could be any type of the aforementioned questions. For these questions, the correct result of the QA system is to indicate that it is unanswerable. SQuADRUn \cite{rajpurkar-etal-2018-know} is an extension of the SQuAD dataset \cite{DBLP:conf/emnlp/RajpurkarZLL16} with over 50,000 unanswerable questions that was introduced to further improve the task of QA.

\subsubsection{Types of Data Sources}
CQA systems can be classified on the basis of the underlying data sources they utilize to find an answer. These underlying data sources could be:

\paragraph{Structured Data Source:}
In a structured document, data is stored in the form of entities. These entities form a separate table. An entity in a table can have multiple attributes associated with it. The definition of these attributes is referred to as the metadata and is stored in a schema. A query language is used to access the data and retrieve relevant information from the schema. Examples of structured data sources are databases and RDF graphs. QALD\footnote{http://qald.aksw.org/} and LC-QuAD \cite{DBLP:conf/semweb/TrivediMDL17} utilize structured data source (i.e., RDF graphs) to answer the questions.

\paragraph{Semi-structured Data Source:}
There is no clearly defined boundary between the stored data and its schema in the semi-structured data sources which makes it quite labor-intensive to build.
An example of a semi-structured data source is XML. The datasets that are designed using semi-structured data sources include TabMCQ \cite{wang-etal-2018-neural-question} and QuaSM \cite{pinto2002quasm}.

\paragraph{Unstructured Data Source:}
There are no pre-defined rules for storing the data in this particular arrangement. The data stored in the unstructured data sources could be of any type and require the use of advanced natural language processing techniques and information retrieval methods to find out the relevant answer. However, the reliability of finding the correct answers is low as compared to the structured data sources. Examples of unstructured datasets are SQuAD \cite{DBLP:conf/emnlp/RajpurkarZLL16}, QuAC \cite{DBLP:conf/emnlp/ChoiHIYYCLZ18}, and CNN \& Daily Mail \cite{DBLP:conf/nips/HermannKGEKSB15}.

\subsubsection{Types of CQA Systems}
Over the past few years, the demand for CQA systems, from both research and commercial perspective, has increased in turn enabling users to search a large-scale knowledge-base (KB) or a text-based corpora written in natural language.  This categorizes the CQA systems into sequential KB-QA agents and conversational machine reading comprehension systems:

\paragraph{Sequential KB-QA:}
KB-QA systems are extremely flexible and easy-to-use in contrast to the traditional SQL-based systems that require users to formulate complex SQL queries \cite{DBLP:journals/pvldb/CuiXWSHW17}. In a real-world scenario, users do not always ask simple questions \cite{DBLP:conf/aaai/SahaPKSC18}. Usually, the questions asked are complex in nature, and therefore, require multi-turn interaction with the KB. Also, once a question has been answered, the user tends to put forward another question that is linked to the previous question-answer pair. This forms the task of sequential QA using knowledge graphs. 

\paragraph{Conversational Machine Reading Comprehension:}
The practical use of text-based QA agents, also referred to as CMRC agents, is more common in the mobile phones than in the search engines (like Google, Bing, and Baidu), wherein concise and direct answers are provided to the users rather than presenting them with a list of possible answers. For instance, if a user intends to look for a popular restaurant in a particular geographical area, the search engine would provide 
her with a search result encompassing options spread on multiple pages, whereas, a CMRC-based dialog agent would ask a few follow-up questions to figure out the preference(s) of the user to subsequently narrow down the search result to one, i.e., possibly the best, answer.  
With the emergence of CMRC, many researchers \cite{DBLP:journals/tacl/ReddyCM19, DBLP:conf/emnlp/ChoiHIYYCLZ18, DBLP:conf/aaai/SahaPKSC18, DBLP:conf/acl/IyyerYC17} have tried inducing a conversational aspect to meet the requirements for the task of CQA by introducing a background context and a series of inter-related questions.

\subsection{What Makes CQA Different from QA?}

\subsubsection{Task-based Differences}
The task of CQA differs from the traditional QA in a number of ways. In traditional QA systems, questions are independent of each other and are based on the given passage. In contrast, questions in CQA are related to each other which poses an entirely different set of challenges including but not limited to:

\begin{scriptsize}
\begin{table}[!tb]
    \centering
    \begin{tabular}{p{0.2\linewidth}  p{0.6\linewidth}}
    \toprule
    \toprule
    \multicolumn{2}{c}{Topic: Staten Island} \\ \hline
      \textbf{Passage:}  & 
Staten Island is one of the five boroughs of New York City in the U.S. state of New York. In the southwest of the city, Staten Island is the southernmost part of both the city and state of New York, with Conference House Park at the southern tip of the island and the state. The borough is separated from New Jersey by the Arthur Kill and the Kill Van Kull, and from the rest of New York by New York Bay. With a 2016 Census-estimated population of 476,015, Staten Island is the least populated of the boroughs but is the third-largest in area at. Staten Island is the only borough of New York with a non-Hispanic White majority. The borough is coextensive with Richmond County, and until 1975 was the Borough of Richmond. Its flag was later changed to reflect this. Staten Island has been sometimes called ``the forgotten borough" by inhabitants who feel neglected by the city government. \\ \hline 
     \begin{tabular}[c]{@{}l@{}}\textbf{Question 1:}\\ \textbf{Answer 1:}\end{tabular} & \begin{tabular}[c]{@{}l@{}}How many burroughs are there?\\ Five.\end{tabular}                                 \\ \hline
\begin{tabular}[c]{@{}l@{}}\textbf{Question 2:}\\ \textbf{Answer 2:}\end{tabular} & \begin{tabular}[c]{@{}l@{}}In what city?\\ New York City.\end{tabular}                                        \\ \hline
\begin{tabular}[c]{@{}l@{}}\textbf{Question 3:}\\ \textbf{Answer 3:}\end{tabular} & \begin{tabular}[c]{@{}l@{}}And state?\\ New York.\end{tabular}                                      \\       \hline   
\begin{tabular}[c]{@{}l@{}}\textbf{Question 4:}\\ \textbf{Answer 4:}\end{tabular} & \begin{tabular}[c]{@{}l@{}}Is Staten island one?\\ Yes. \end{tabular}                                      \\       \hline  
\begin{tabular}[c]{@{}l@{}}\textbf{Question 5:}\\ \textbf{Answer 5:}\end{tabular} & \begin{tabular}[c]{@{}l@{}} Where is it? \\ In the southwest of the city\end{tabular}                                      \\       \hline  
\begin{tabular}[c]{@{}l@{}}\textbf{Question 6:}\\ \textbf{Answer 6:}\end{tabular} & \begin{tabular}[c]{@{}l@{}}What is it sometimes called?\\ The forgotten bourough.\end{tabular}  \\  \hline            
\begin{tabular}[c]{@{}l@{}}\textbf{Question 7:}\\ \textbf{Answer 7:}\end{tabular} & \begin{tabular}[c]{@{}l@{}}Why?\\ Because the inhabitants feel neglected by the city \\ government.\end{tabular} \\                          \\ \toprule \toprule
\hline
    \end{tabular}
    \caption{A chunk of a dialog from the
    CoQA dataset \cite{DBLP:journals/tacl/ReddyCM19}}
    \label{coqadataset}
\end{table}
\end{scriptsize}

\begin{itemize}
    \item In order to find the correct answer for the question at hand, the model needs to encode not only the current question and source paragraph, but also the previous history turns. More specifically, as shown in Table~\ref{coqadataset}, Question 2 and Question 3 are related to Question 1.
    
    \item The turns in CQA are of different nature. Some questions require more detailed information (i.e., \textit{drilling down}), some may require information about some topic previously discussed (i.e., \textit{topic shift}), some may ask about a topic again after it had been discussed (i.e., \textit{topic return}), and some questions may ask for the clarification of topic (i.e., \textit{clarification question}) \cite{DBLP:conf/naacl/Yatskar19}. All of these characteristics are incremental in nature and present challenges that most of the top-performing QA models fail to address directly, such as pragmatic reasoning and referring back to the previous context applying co-reference resolution. In Table~\ref{coqadataset}, Question 2 is an example of a drill down question, Question 7 is a clarification question and ``\textit{it}'' in Question 5 ``\textit{where is it?}'' requires co-reference resolution.
\end{itemize}

\subsubsection{Architectural Differences}
The architecture of a CQA model is similar to the one of a QA system on the base level. However, to introduce the conversational touch to the system, a CQA model extends the traditional QA system by introducing a few modules:
\begin{itemize}
    \item A traditional single-turn KB-QA system encompasses a semantic parser and a knowledge base reasoning (KBR) engine. In addition to these, a sequential KB-QA system encompasses a dialog manager, which is responsible for tracking the previous dialog states and determines what question to ask next to help a user search the KB effectively 
    for 
    an answer.
    \item A CMRC system differs from a traditional MRC system in two aspects. First, the encoder is embedded with a sub module referred to as history modeling module, which is responsible for not only encoding the current question and the given passage, but also the history turns of the conversation. Second, a reasoning module is extended to generate an answer, that might not be directly given in the passage, using pragmatic reasoning \cite{DBLP:journals/tacl/ReddyCM19}. 
\end{itemize}

It is worth noting here that the paradigm of CQA is an emerging one, 
which is not well studied in contrast to traditional QA systems. Therefore, not many research papers are available.
The architecture and researches carried out in both sequential KB-QA systems and CMRC systems 
will be 
discussed in detail in Section~\ref{seq kb-qa} and Section~\ref{cmrc}.

\section{Sequential KB-QA Systems}
\label{seq kb-qa}
A knowledge base (KB) is a structured information repository used for knowledge
sharing and management purposes \cite{martinez2015automated}. Freebase \cite{bollacker2008freebase}, NELL \cite{mitchell2018never}, DBpedia \cite{lehmann2015dbpedia}, and Wikidata\footnote{https://www.wikidata.org/wiki/Wikidata:Main\_Page} are 
well-known examples of large-scale graph-structured knowledge
bases also termed as the Knowledge Graphs (KGs) and have 
become significant resources when 
dealing with open-domain questions. The KGs are known to be a graphical representations of a KB, and a typical KG comprises of triples encompassing subject, predicate, object triples \textit{(s,r,t)}, wherein \textit{r} is a relation or predicate between the entities \textit{s} and \textit{t} \cite{DBLP:journals/ftir/GaoGL19}. They play an important role in bridging up the lexical gap by providing additional information about relations which in turn helps in gaining more detailed information about the context.  The knowledge graphs have seen their successful applications in various NLP tasks such as text entailment, information retrieval, and QA \cite{Zou_2020}.
\begin{figure} [tb!]
\center
  \includegraphics[scale = 0.35]{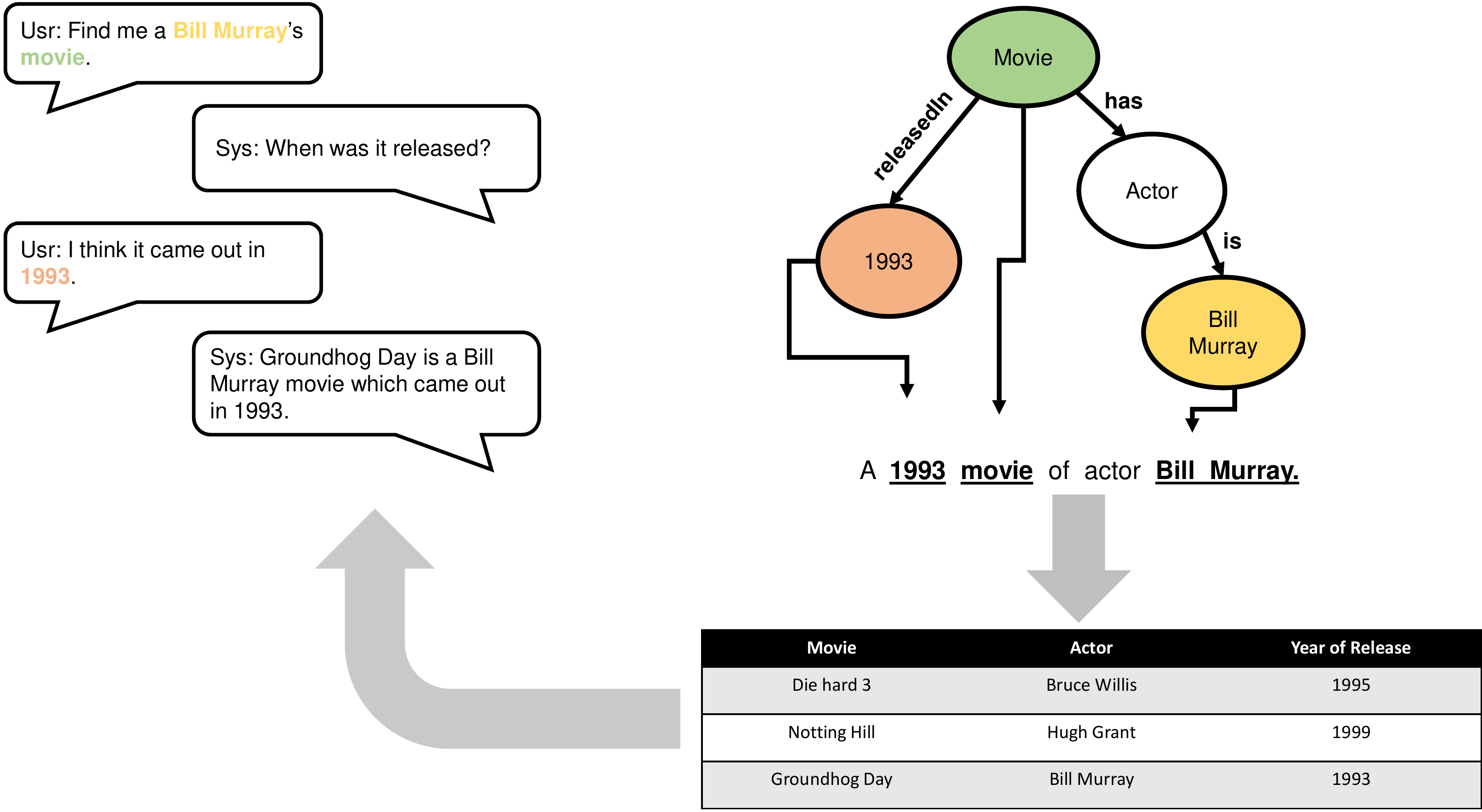}
  \caption{Aligning knowledge and conversation in sequential KB-QA.} 
  \label{movie}
\end{figure}
The task of 
QA 
over large-scale KB-QA systems has seen its progress from simple single-fact task to complex queries requiring multi-hop interaction and traversal of the knowledge graphs. These come under the category of single-turn QA where a user puts forward a question and the system finds the best possible answer for it. Though KB-QA based agents improved the flexibility of QA process to a considerable extent, nevertheless, it is irrational to believe that these systems could constitute complex queries without having complete knowledge about the organizational structure of the KB to be questioned \cite{DBLP:conf/nips/GuoTDZY18}. Thus, sequential KB-QA system is a more optimal option as it lets the users query the KB interactively.

The interactive sequential KB-QA system is useful in many commercial areas such as making a restaurant reservation \cite{sun2020multi}, finding a hotel 
in a new city, finding a movie-on-demand \cite{DBLP:conf/acl/DhingraLLGCAD17}, or asking for relevant information based on certain attributes. Fig.~\ref{movie} illustrates how a sequential KB-QA system aims to find a movie based on specified attributes by a user. If it is a traditional KB-QA system, the conversation would have ended after the first turn with a number of results. But under the sequential KB-QA setting, the system asks the follow-up questions 
for the specific details about the current question and present the user with the most appropriate answer.
\begin{figure} [tb!]
\center
  \includegraphics[width=10cm, height=5cm]{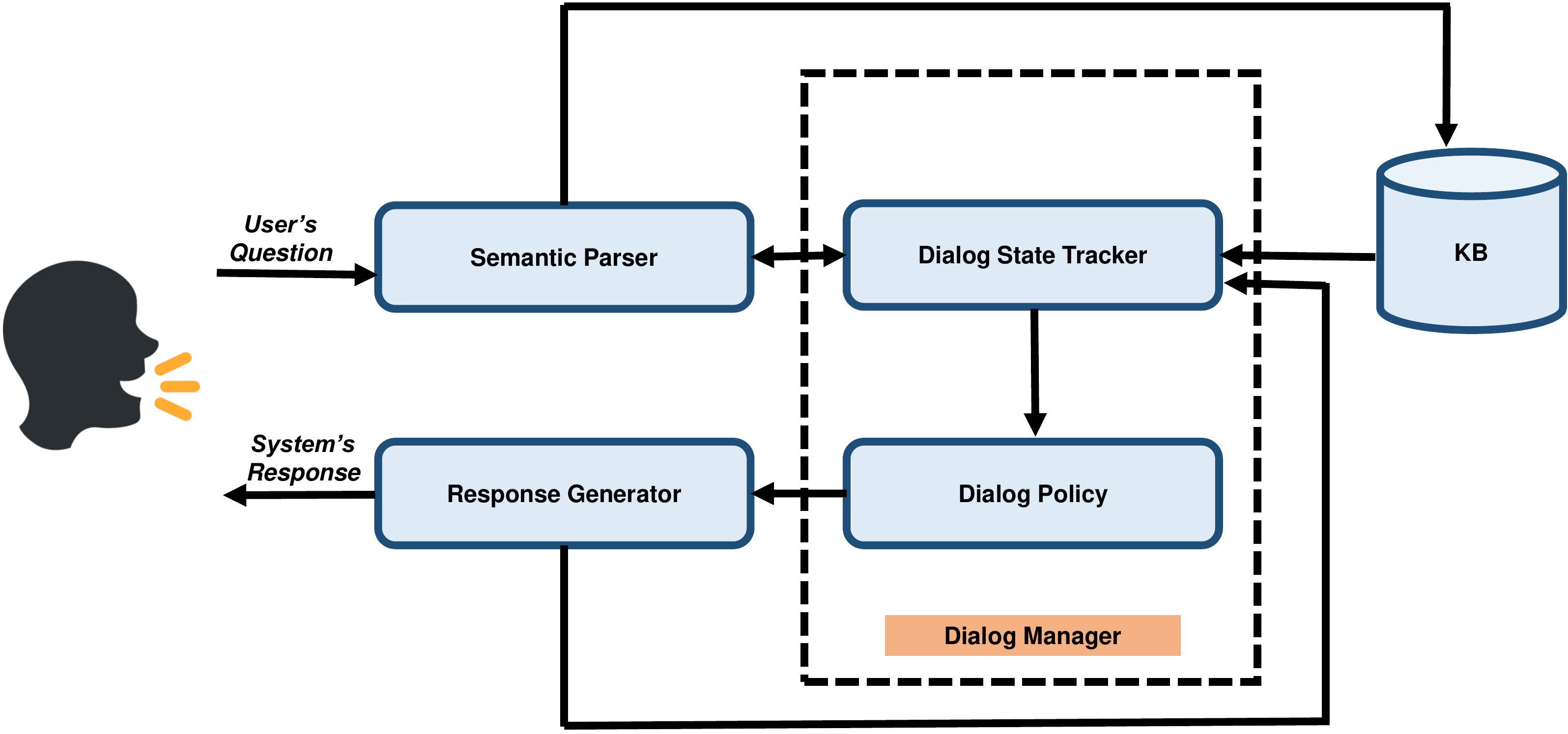}
  \caption{A high level diagram of sequential KB-QA.} 
  \label{kbqagen}
\end{figure}

The core architecture of a sequential KB-QA system comprises of a semantic parser and an inference engine, along with the addition of a dialog manager, that keeps track of the previous turns and decides which questions to ask to help the user query the KB effectively. The high-level 
architecture of a sequential KB-QA is depicted in Fig.~\ref{kbqagen}, which
%
consists of: 
i) \textit{Semantic Parser}, ii) \textit{Dialog Manager}, and iii) \textit{Response Generator}. The semantic parser is responsible for mapping input along with the previous context into a semantic representation (logical form) to query the KB. The dialog manager keeps track of the dialog history (i.e., QA pairs and DB state) and updates it accordingly \cite{suhr-etal-2018-learning}. It is also responsible for selecting the system's next action (i.e., to provide an answer or to ask a clarification question) based on the current question using dialog policy. The process of dialog policy can be either trained on dialogs  \cite{wen-etal-2017-network, DBLP:conf/acl/DhingraLLGCAD17} or programmed \cite{wu2015probabilistic}. 
At the end, 
the response generator converts the system's action into natural language response. However, certain new approaches \cite{DBLP:conf/emnlp/MullerPSNA19,DBLP:conf/cikm/ChristmannRASW19} are working towards the elimination of the semantic parser module as it requires extensive and expensive labeling of data.

\subsection{Semantic Parser}
The notion of semantic parsing can be thought of as a process of mapping natural language text into meaningful logical forms and has emerged as a significant technical component for designing KB-QA systems \cite{cheng-etal-2019-learning}. Once a correct logical form has been obtained, it can be executed on the knowledge source in the form of a query to obtain answer denotations.

Iyyer et al. \cite{DBLP:conf/acl/IyyerYC17} introduced the task of semantic parsing for sequential QA by creating a dataset of simple inter-related questions out of a complicated WikiTableQuestions dataset \cite{pasupat-liang-2015-compositional}. The proposed model, called Dynamic Neural Semantic Parsing (DynSP), is a weakly supervised structured-output learning approach based on the reward-guided search. Given a question along with a table, the model forms a semantic parsing problem as a state-action search problem wherein each state denotes a partial or complete parse and each action can be considered as an operation to extend the parse. Unlike traditional parsers, DynSP explores and constructs different neural network structures for different questions.

The aforementioned approach maps only the current utterance into its logical form which makes it difficult for the system to interpret the meaning of the utterance especially where co-reference resolution is required. To address this shortcoming, the authors in \cite{DBLP:conf/nips/GuoTDZY18} proposed Dialog-to-Action (D2A) to facilitate the use of previous utterances (both questions and answers) concatenated with the current question. The task of generating logical form can be regarded as the prediction of a series of actions, and each of 
them corresponds to simple grammar rules. 

However, the model of D2A suffers from the problem of error propagation as it learns to reproduce previously generated actions, which might be incorrect. To overcome the issue of error propagation and ambiguous entity-linking, the step-wise framework is improved by multi-task learning for sequential KB-QA systems \cite{DBLP:conf/emnlp/ShenGQGTDLJ19}. This model, Multi-task Semantic Parsing (MaSP), learns pointer-based semantic parsing and entity-detection simultaneously as they are closely related. 
The joint learning could enhance 
 the performance of the CQA task. 
 Specifically, 
 the input consists of the current question and historical interactions are passed through an encoder based on Transformer \cite{NIPS2017_3f5ee243} to generate the context-aware embeddings. The model employs pointer network \cite{NIPS2015_29921001} to locate the targeted entity and a number in the given question. The use of the pointer network comes with two advantages: i) it handles the co-reference resolution by learning the context of the entity, and ii) it reduces the size of decoding vocabulary significantly from several million to several dozen. 
 The model also incorporates a type-aware entity detection module in which the prediction is fulfilled in joint space of IOB (inside, outside, beginning) tagging and corresponding entity type for disambiguation. In the end, grammar guided decoder is used to infer logical forms that can be executed on the KB.

 The model of MaSP suffers from the issue of producing ambiguous results because the task of jointly learning predicate and entity classification share no common information except for the supervision signals propagated to the classifiers. The issue was overcome by another recently introduced model called mu\textbf{L}ti-task sem\textbf{A}ntic par\textbf{S}ing with tr\textbf{A}nsformer and \textbf{G}raph atte\textbf{N}tion n\textbf{E}tworks (LASAGNE) \cite{kacupaj-etal-2021-conversational}. The model performs multi-task learning by utilizing a Transformer \cite{NIPS2017_3f5ee243} supplemented with a Graph Attention Network (GAT) \cite{DBLP:conf/iclr/VelickovicCCRLB18}. The model uses a Transformer to generate the logical forms of a natural language question, while GAT model is utilized to exploit the correlations between predicate and entity types due to its message-passing ability between the nodes. The authors also proposed an entity recognition module that contributes in detecting, linking, filtering, and permuting all the relevant entities in the generated logical forms. Unlike MaSP, LASAGNE use both sources of information, the encoder and the entity recognition module to perform these operations which makes the process of re-learning entity information from the context of the current question avoidable.
 
\subsection{Dialog Manager}
Conversational history plays a significant role when generating the logical forms of natural language utterances. Once a logical form is obtained, the system is in a better state to decide its next action, i.e., to ask a clarification question or provide an answer to a question.  

Dialog-to-action \cite{DBLP:conf/nips/GuoTDZY18} incorporates a dialog memory to store the historical interaction of a user. The model consists of a bidirectional RNN with a Gated Recurrent Unit (GRU) \cite{69e088c8129341ac89810907fe6b1bfe} as an encoder to convert the input (previous question answer pairs concatenated with the current question) into a sequence of context vector. A grammar-guided decoder (GRU with attention mechanism) generates an action sequence based on the context vector \cite{luong-etal-2015-effective}. The dialog memory used in the model encompasses entities, predicates, and action sub-sequences which could be replicated selectively as decoding proceeds.  

LASAGNE \cite{kacupaj-etal-2021-conversational} incorporate the dialog history based on previous interactions as an additional input to the model for handling ellipsis and coreference. The final input consists of the previous question-answer pair and the current question. The utterances are separated using a \textit{[SEP]} token and at the end of the last utterance, a context token \textit{[CTX]} is appended. The conversation is tokenized using WordPiece tokenization \cite{DBLP:journals/corr/WuSCLNMKCGMKSJL16} and then pre-trained GloVe model \cite{pennington-etal-2014-glove} is used to embed the words into vector representations.

\subsection{Response Generator}
In NLP tasks, response generation is the last and vital step to generate system utterances for a user, and the introduction of pre-trained language models has been a game-changing factor for the promising field of language generation over the past few years. Peng et al. \cite{DBLP:conf/emnlp/PengZLLLZG20} introduced a model based on Open AI's Generative Pre-training (GPT) \cite{radford2018improving} called Semantically-Conditioned Generative Pre-training (SC-GPT). The paper introduces a dataset called FewshotWOZ\footnote{https://github.com/pengbaolin/SC-GPT} to simulate the process of few-shot learning for limited data labels. SC-GPT generates semantically controlled responses and is trained in three steps: i) initially, it is pre-trained on massive plain corpora so that it can better generalize to new domains, ii) further pre-training is 
conducted on dialog-act specific huge corpora to gain the capability of controllable generation and finally, and 
iii) a limited amount of domain labels are used to fine-tune the model for its adaptation to the target domain. 

Another framework called NLG-LM \cite{zhu2019multi} employs multi-task learning to not only generate semantically correct responses, but also maintain the naturalness of the conversation. The model utilizes sequence-to-sequence architecture to simultaneously train the Natural Language Generation (NLG) and Language Modeling (LM) tasks. The language modeling task, carried out in decoder, is incorporated on human generated utterances to bring out more language-related elements. In addition to that, the unsupervised nature of the language model eliminates the need for a massive amount of unlabelled data for training purposes.

\subsection{Sequential KB-QA Approaches without Semantic Parser}
There exists extensive research work in semantic parsing, wherein deep neural networks have been utilized for training models in a supervised learning setup over manually generated logical forms. However, generating labeled data for this task could be exhaustive and expensive \cite{DBLP:conf/emnlp/MullerPSNA19}. To address this issue, a new research direction has been recently investigated that utilizes weak training for semantic parsing where training data consists of question and answers and the structured resources are used to restore the logical representations that would result in the right answer. 

In \cite{DBLP:conf/aaai/SahaPKSC18}, the authors proposed a model which is an amalgamation between Hierarchical Recurrent Encoder-Decoder (HRED) \cite{serban2016building} model and key-value memory network \cite{miller-etal-2016-key} to present the fusion of dialog and QA process. 
HRED is responsible for generating high-level and low-level representations of an utterance and the context. Candidate tuples are selected in which the entity appears as subject/object. These candidates i.e., tuples are stored in a key-value memory network as key-value pair, where the key contains the relation-subject pair and the value contains the embeddings of the object. The model makes multiple passes (turns) to attend to different aspects of the question especially in the case of complex questions. A decoder is used to generate answer sequences.

\afterpage{
\begin{landscape}

\scriptsize
\RaggedLeft
\begin{longtable}{|p{1.5cm}|p{3.5cm}|p{3.0cm}|p{4.0cm}|p{4.0cm}|}
 \caption{Recent studies on sequential KB-QA (2016-2021).} \label{tab:long} \\

\hline \multicolumn{1}{|c|}{\textbf{Ref.}}  &  \multicolumn{1}{|c|}{\textbf{Contribution(s)}} &  \multicolumn{1}{|c|}{\textbf{Techniques Used}} & \multicolumn{1}{|c} {\textbf {Merits}} & \multicolumn{1}{|c|}{\textbf{Demerits}} \\ \hline 
\endfirsthead

\multicolumn{5}{c}%
{{\bfseries \tablename\ \thetable{} -- continued from previous page}} \\
\hline \multicolumn{1}{|c|}{\textbf{Ref.}}  &  \multicolumn{1}{|c|}{\textbf{Contribution(s)}} &  \multicolumn{1}{|c|}{\textbf{Techniques Used}} & \multicolumn{1}{|c} {\textbf {Merits}} & \multicolumn{1}{|c|}{\textbf{Demerits}} \\ \hline 
\endhead

\hline \multicolumn{5}{|r|}{{Continued on next page}} \\ \hline
\endfoot

\hline \hline
\endlastfoot 
DynSP \color{blue}$\textbf{\cite{DBLP:conf/acl/IyyerYC17}}$ &  Introduced the task of semantic parsing for sequential KB-QA.
        &  Dynamic Neural Network structure.&
        \begin{tableitems}[nosep,after=\strut]
                    
                         \item Reward-guided search reduces the number of queries to be labelled.
                         
        \end{tableitems} 
                 &\begin{tableitems}[nosep,after=\strut]
                    
                         \item The parse language is not comprehensive enough to represent the semantic parses of the sentences in dataset.
                         \item The table based search-space approach cannot be scaled up to cater the needs of large-sacle curated KGs.
        \end{tableitems} \\ \hline
        HRED + KVmem \color{blue}$\textbf{\cite{DBLP:conf/aaai/SahaPKSC18}}$ & 
      Introduced sequential KB-QA dataset consisting of complex questions. 
        & End-to-end model based on HRED and KV memnet.
        & \begin{tableitems}[nosep,after=\strut]
                    
                         \item Incorporates dialog history with the current utterance. 
                         \item Works well with simple and direct questions. 
        \end{tableitems}  &\begin{tableitems}[nosep,after=\strut]
                    
                         \item Performs poorly with complex questions.
                         \item Doesn't work well with indirect or incomplete questions.
                         \item KV memnet has flat organization of story which makes it unsuitable for complex questions.
        \end{tableitems}  \\ \hline
         D2A \color{blue}$\textbf{\cite{DBLP:conf/nips/GuoTDZY18}}$ & 
     Introduced history interaction as a part of input to deal with enormous ellipsis phenomena. 
        & A bidirectional RNN with a GRU is used as an encoder. A grammar guided decoder along with a dialog memory component is used to generate action sequences. 
        & \begin{tableitems}[nosep,after=\strut]
                    
                         \item The model can effectively handle the contextual references  
                         \item The parser introduced is capable of parsing various types of question.
        \end{tableitems}  & \begin{tableitems}[nosep,after=\strut]
                    
                         \item Error propagation may occur because the model replicates previously generated action-sequences which might be incorrect.
                         \item The supervision signals cannot be shared among the model for mutual benefits as they are learned independently for the subtasks.
                         \item Ambiguous entity linking.
        \end{tableitems}\\ 
        \hline
         MaSP \color{blue}$\textbf{\cite{DBLP:conf/emnlp/ShenGQGTDLJ19}}$ & 
     Multi-task learning for sequential KB-QA.
        & Utilizes Transformer as a contextual encoder and a pointer-equipped decoder.
        & \begin{tableitems}[nosep,after=\strut]
                    
                         \item Reduces the risk of error propagation by jointly learning semantic parsing and entity detection.
                         \item Works well with co-reference resolution.
                         \item Addresses ambiguous entity-linking by leveraging contextual features of the input.
        \end{tableitems}  &\begin{tableitems}[nosep,after=\strut]
                    
                         \item May result in spurious logical form.
                         
        \end{tableitems}\\ 
         \hline
         LASAGNE \color{blue}$\textbf{\cite{kacupaj-etal-2021-conversational}}$ & 
         
     Improved multi-task semantic parsing for sequential KB-QA.
        & 
        \begin{tableitems}[nosep,after=\strut]
        \item Utilizes Transformer model to generate logical forms, while the graph attention model is used to exploit correlations between entity type and predicates.
        \item Introduced an entity detection module which detects, links, and permutes all the relevant entities.
        \end{tableitems} 
        & \begin{tableitems}[nosep,after=\strut]
                    
                         \item Eliminates the risk of producing ambiguous results by sharing signals between entity and predicate nodes.
                         \item Works well with co-reference resolution.
                         \item Improves the process of entity detection and linking by utilizing information from both entity detection module and encoder.
        \end{tableitems}  &\begin{tableitems}[nosep,after=\strut]
                    
                         \item May result in spurious logical forms which affects the model's performance in answering clarification and ellipsis-based questions.
                         
        \end{tableitems}\\ \hline
        GNN + PointerNet \color{blue}$\textbf{\cite{DBLP:conf/emnlp/MullerPSNA19}}$ & Conversation processing around structured data.
        & Neural approach based on GNN and pointer network.
        & \begin{tableitems}[nosep,after=\strut]
                        \item Eliminates the need of semantic parsing.
                        \item Handles conversational context stored in tables, effectively.
        \end{tableitems}  & \begin{tableitems}[nosep,after=\strut]
                        \item Table-search methods cannot scale to large real-world KGs involving qualitative or logical comparison.
                    
        \end{tableitems} \\ \hline
       CONVEX \color{blue}$\textbf{\cite{DBLP:conf/cikm/ChristmannRASW19}}$ & 
      \textit Completion of incomplete follow-up questions.
        & Symbolic approach.
        & \begin{tableitems}[nosep,after=\strut]
                    
                         \item Automatically infers missing or ambiguous pieces for follow-up questions.
                         \item Eliminates the need for intermediary representation of the context and given question 
        \end{tableitems}  & \begin{tableitems}[nosep,after=\strut]
                    
                         \item May result in combinatorial explosion if sub-graphs not expended carefully. 
        \end{tableitems}   \\

\hline 
\end{longtable}
\end{landscape}
}

Another approach in \cite{DBLP:conf/emnlp/MullerPSNA19} presents a table-centered sequential KB-QA model which, instead of learning the intermediate learning forms, encodes the structured resources (i.e., tables) along with the questions and answers from the conversational context. The approach encodes tables as graphs by representing cells, columns, and rows.  The column represents the main features of the questions and cells contains the relevant values. To handle the follow-up questions, the model adds previous answers by marking all the columns, rows, and cells with nominal features. It uses a Graph Neural Network (GNN) \cite{scarselli2008graph} based encoder to encode the graph by generating vector representation of the edge label between the two nodes. The copy mechanism based on the pointer network, instead of selecting symbols from output vocabulary, then predicts the sequences of answer rows and columns from the given input.

CONVEX (CONVersational KB-QA with context EXpansion) \cite{DBLP:conf/cikm/ChristmannRASW19} employs unsupervised method to answer sequential questions (follow-up questions) by keeping track of the conversational context using predicates and entities appeared so far. The initial question is used for initializing and selecting a small sub-graph of the knowledge graph. The essence of this approach is the graph exploration algorithm that tends to expand a frontier aptly to find the possible candidate answers for the follow-up questions. The right answer from the candidate answers is selected by calculating weighted proximity. The top-scoring answer (in the range of 0 to 1) will be returned as the answer to the current question. 

Table~\ref{tab:long} 
summarizes the major contributions, the techniques exploited, and the merits and demerits of the aforementioned approaches.

\section{Conversational Machine Reading Comprehension}
\label{cmrc}
Most of the work carried out in the field of machine reading comprehension is based on single-turn QA which is unlikely in the real-world scenario since humans tend to seek information in a conversational context \cite{DBLP:conf/emnlp/RenXCY18}. For instance, a user might ask, ``\textit{Who is Christopher Columbus?}'' and based on the answer received, he might further investigate, ``\textit{Where was he born?}'' and ``\textit{What was he famous for?}''. It is easy for a human to decipher that here ``\textit{he}'' in the follow-up questions refer to ``\textit{Christopher Columbus}'' from the first question. But when it comes to a machine to comprehend the context, it poses a set of challenges such as co-reference resolution or conversational history \cite{liu2019neural}, which most of the state-of-the-art QA systems do not address directly.

\begin{figure} [tb!]
\center
  \includegraphics[width=8cm, height=5cm]{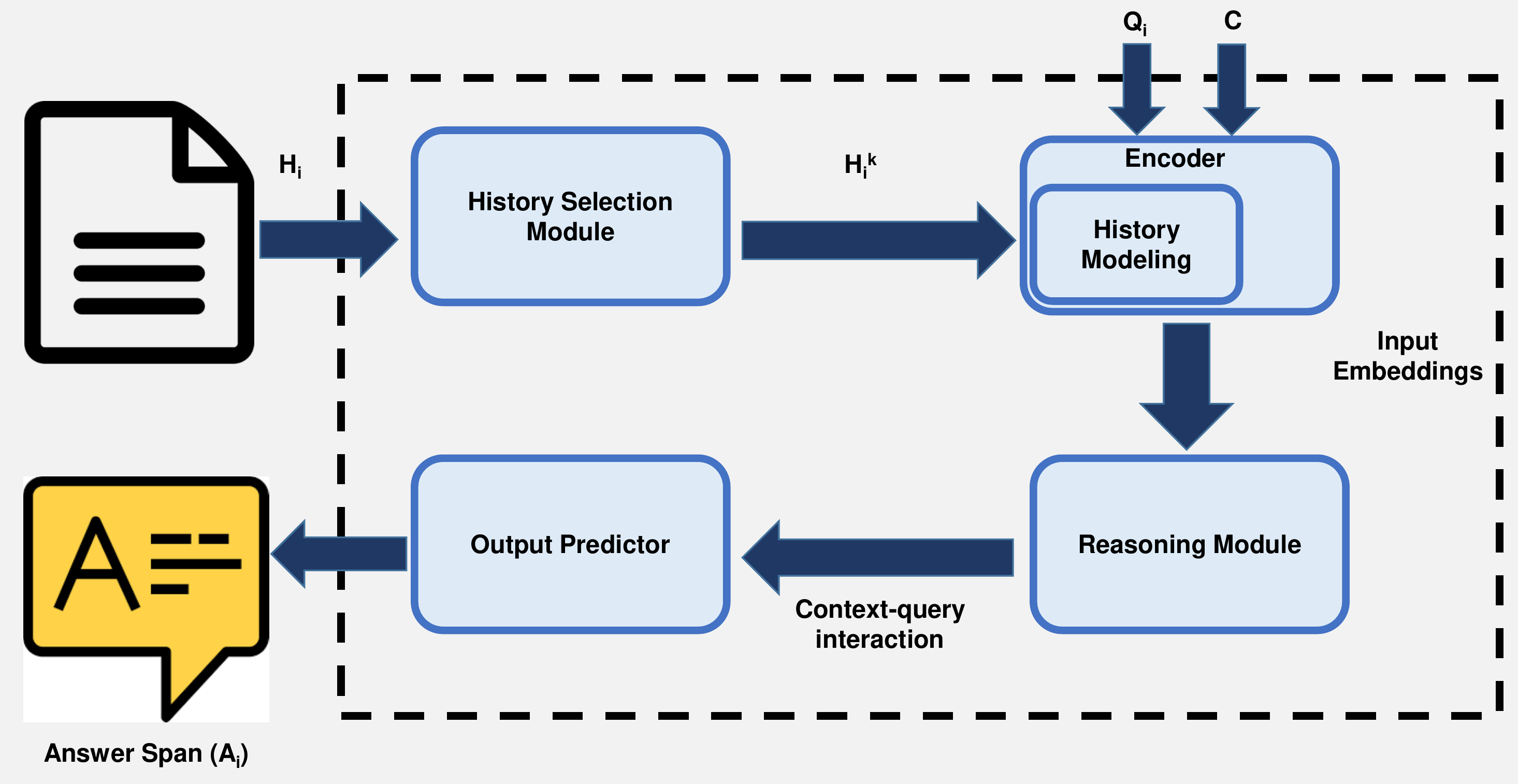}
  \caption{Generic framework of a CMRC model which consists of i) history selection module that selects ${H_i}'$ history turns from the conversational history context $H_i$, ii) encoder that transforms the tokens of ${H_i}', C, Q_i$ into input embeddings, iii) reasoning module is responsible for performing contextual integration of input embeddings into contextualized embeddings to perform reasoning, and iv) output predictor predicts the answer $A_i$ on the basis of context-query interaction.} 
  \label{architecture}
\end{figure}

A typical MRC model consists of  three main functions namely: i) encoding the given context and question into a  set of symbolic representations called embeddings in a neural space, ii) reasoning through the embeddings to find out the answer vector in the neural space, and iii) decoding the answer vector to produce natural language output \cite{DBLP:journals/ftir/GaoGL19}. In \cite{DBLP:conf/sigir/Qu0QCZI19}, the authors proposed a modification by introducing two modules, i.e., history selection and history modeling modules to address the aforementioned challenges to incorporate the conversational aspect, hence introducing the task of CMRC. 
%
%
 Formally, given a context $C$, the conversation history in the form of question-answer pairs $Q_1, A_1, Q_2, A_2,...,Q_{i-1}, A_{i-1}$, and a question $Q_i$, the CMRC model needs to predict the answer $A_i$. The answer $A_i$ can either be a free-form text with evidence \cite{DBLP:journals/tacl/ReddyCM19} or a text span \cite{DBLP:conf/emnlp/ChoiHIYYCLZ18}.
%
%
The flow of a general CMRC model is depicted in Fig.~\ref{architecture}. 

We will discuss these modules separately in the 
rest of this section, along with the techniques and trends utilized in each of them for the successful design and implementation of a CMRC model.

\subsection{History Selection Module}
To enable the CMRC model to predict the answer span more accurately, it is necessary to introduce the previous context along with the source passage and current question. However, context utterances that are relevant to the query are useful, whereas, the irrelevant ones may bring 
more noise \cite{tian-etal-2017-make, 10.1007/978-981-16-0010-4_5}. Thus, the careful selection of conversational history turns is quite critical for the model. History selection process can be categorized as: 
\paragraph{Selecting \textit{K} turns.} SDNET \cite{DBLP:journals/corr/abs-1812-03593} BIDAF++ \cite{DBLP:conf/emnlp/ChoiHIYYCLZ18}, ORConvQA \cite{qu2020open}, and WS-OR-ConvQA \cite{DBLP:conf/ecir/QuYCCKI21} utilize conversation history by incorporating \textit{K} rounds of history turns.

\paragraph{Immediate History Turns.} BERT with 2-ctx \cite{ohsugi-etal-2019-simple} suggests that incorporating immediate two turns can be helpful in predicting the right answer span, whereas, BERT-HAE \cite{DBLP:conf/sigir/Qu0QCZI19} claims that incorporating 5-6 conversational history turns contributes more in finding the correct answer span. However, both models demonstrate a dramatic degradation in the performance with the increase in the number of history turns.

\paragraph{Dynamic History Selection.} 
In \cite{DBLP:conf/naacl/Yatskar19}, the authors pointed out that the dialog features like topic return or topic shift may not align with the concept of selecting immediate dialog turns. Therefore, in order to address this shortcoming, History Answer Modeling (HAM) \cite{DBLP:conf/cikm/QuYQZCCI19} was introduced as a dynamic policy that weighs the previous dialog turns on the basis of their contribution to answering the current question. The model assigns weight by attending the previous history turns at a token level or sentence level and combine the same with the current turn’s representation. 

Another approach, referred to as Env-ConvQA \cite{qiu2021reinforced}, proposed a dynamic \textit{k}-history turns selection process based on reward-based reinforced backtracking policy. The model treats the process of extracting the relevant history turns as a sequential decision making process. The model acts on the provided history turns and backtracks through each turn one by one to decide whether the turn is relevant to the current question or not. 

\subsection{Encoder}
This component is responsible for converting the tokens of the source passage, current question, and the selected history turns into fixed-length vectors which are subsequently provided as an input to the reasoning module. Although the internals of an encoder may vary from approach to approach depending on the input required by the reasoning module, nevertheless, the high-level encoding generally involves transforming and combining different context-dependent word embeddings, including but not limited to, ELMo \cite{peters-etal-2018-deep}, GloVE \cite{pennington-etal-2014-glove}, and BERT \cite{DBLP:conf/naacl/DevlinCLT19}. To improve the impact of these embeddings, additional features such as 
Parts of Speech (POS) tags and History Answer Embeddings (HAE) have also been incorporated as a part of the input.  These embeddings can be categorized into conventional word embeddings and contextualized word embeddings.

\paragraph{Conventional Word Embeddings.}
 This technique is responsible for encoding of words into low-dimensional vectors. The encoding is done in such a way that the inter-related tokens are placed in close proximity to each other in vector space to make the identification of co-relation easy between them. Several methods for generating distributed word representation have been proposed in the literature, with the most popular and efficient being GloVE \cite{pennington-etal-2014-glove} and Word2Vec \cite{DBLP:journals/corr/abs-1301-3781}. However, these methods fail to determine the accurate meaning of the words with respect to their given context.
 
\paragraph{Pre-trained Contextualized Word Embeddings.}
Though the conventional word embeddings method yields good results in identifying and establishing the correlation between the words encoded in low-dimensional vectors, 
they still fail to capture the contextual representations sufficiently. To be accurate, the distributed word representations generated for a single word are the same in varying contexts. To overcome this issue, the idea of contextualized embeddings was put forward by the researchers. These embeddings are pre-trained on large corpora of text and are then utilized as either distributed word embeddings or are fine-tuned according to the specific downstream task. This comes under the category of transfer learning and has obtained astonishing results in various NLP-based tasks \cite{DBLP:conf/naacl/DevlinCLT19, peters-etal-2018-deep, NEURIPS2019_dc6a7e65}. 

The most successful application of these embeddings has been in the field of machine comprehension. One of the very first in the series is Context Vectors (CoVe) \cite{NIPS2017_7209} which utilizes Seq2Seq models \cite{NIPS2014_a14ac55a} to train Long Short-Term Memory (LSTM) \cite{hochreiter1997long} encoders on a large scale dataset. The encoder then utilizes the obtained results on other downstream NLP tasks. Proposed by \cite{peters-etal-2018-deep}, Embeddings from Language Model (ELMo) is a successor of CoVE and embeddings are obtained by training a bi-directional Language Model (biLM). These embeddings can generate more accurate representations of the words as instead of using the results from the topmost layer of biLM, it combines outcomes from all the layers of biLM into one vector and assign a weighting score that is task-specific. Another popular model in terms of language understanding is Transformer \cite{NIPS2017_3f5ee243} which is a sequence transduction model based on multi-headed attention, thus entirely eliminating the need of utilizing multiple recurrent layers that are part of the most encoder-decoder architectures. This mechanism of self-attention makes the transformer model more efficient and parallelizable in learning the context of the input sequences. The most recent and top-trending one in the series is BERT \cite{DBLP:conf/naacl/DevlinCLT19} that has addressed the issue of unidirectionality used in training of different language models such as Generative Pre-training (GPT) \cite{radford2018improving} and GPT-2 \cite{radford2019language}. Due to the bi-directional property and the powerful 
transformer \cite{NIPS2017_3f5ee243} architecture, BERT’s performance exceeds the top-performing models in many NLP downstream tasks \cite{DBLP:conf/naacl/DevlinCLT19}.

\subsection{History Modeling}
The process of history modeling is generally carried out in the encoder module where the conversation history is integrated with the context and current question to form a complete input. 

We describe it as a separate module for easy understanding and better readability. 
Different models employ different techniques or a combination of these to introduce conversational history turns as a part of an input. A brief description of each of these techniques is given as follows:

\paragraph{Appending the Conversation History.}
One of the most common ways to include the selected history turns (previous question answer pairs) as a part of the input is by appending them with the current question \cite{qu2020open, DBLP:journals/corr/abs-1812-03593,qiu2021reinforced} . This is further modified at sublevel by some approaches \cite{DBLP:conf/ijcai/0022WZ20, ohsugi-etal-2019-simple} via appending only history questions along with the turn number encoded with it. In \cite{DBLP:conf/emnlp/ChoiHIYYCLZ18}, the authors claimed that adding dialog-turn in the input yields better results practically.

\paragraph{Introducing History Answer Markers in the Given Context.}
Another trend seen recently in modeling the conversation history is encoding the context tokens in history answer embeddings markers \cite{DBLP:conf/sigir/Qu0QCZI19}. The advantage of using these tokens is that they work as an indicator to point out whether a context token is a part of history answer or not. Another variation of HAE is Positional HAE (POS-HAE) \cite{DBLP:conf/cikm/QuYQZCCI19}, wherein position information of dialog turn relative to the current question is also encoded. This enables the model to capture the spatial patterns of history answers in context. 

\paragraph{Generating Latent Representations using Context Tokens.}
One of the attributes of successful CMRC models is being able to grasp the flow of the conversation. Since the flow of the conversation based on the given context, it can be captured by generating latent or intermediate representations of the context tokens rather than using the raw inputs. Such approaches \cite{DBLP:conf/iclr/HuangCY19, DBLP:conf/acl-mrqa/YehC19} fall under the category of flow-based methods. 

\subsection{Reasoning Module}
CMRC models can be grouped based on how they perform the process of reasoning. For \textit{single-step reasoning}, 
the model passes the contextualized input (context, question, and history turns) only across one layer and generates the answer. In contrast, for \textit{multi-step reasoning},
the contextualized input is fused across multiple layers to produce history-aware contextualized output embeddings. Generally, the input for this module consists of multiple sequence sets which are then fused in multiple layers and are usually interwined with an attention mechanism to generate accurate output embeddings. On the basis of underlying techniques, the reasoning process can be categorized as
\textit{conventional methods}, \textit{pre-trained language models}, \textit{flow-based models}, and \textit{open-retrieval based models}.

\subsubsection{Conventional Methods}
Several sequence models employing different mechanisms like self-attention and bidirectional attention are a common choice for carrying out the task of conversational machine comprehension. 
%
Famous as CoQA's baseline, DrQA+PGNet \cite{DBLP:journals/tacl/ReddyCM19} leverages the strengths of two powerful models, i.e., Pointer-Generator Network (PGNet) \cite{see-etal-2017-get} and Document Reader (DrQA) \cite{chen-etal-2017-reading}. DrQA, based on bi-directional LSTM (biLSTM), first provides cues from the answer evidence in the given context. PGNet, which utilizes an attention-based Seq2Seq model \cite{DBLP:journals/corr/BahdanauCB14}, decodes the found evidence to predict the final answer. 
 
BiDAF++ \cite{DBLP:conf/emnlp/ChoiHIYYCLZ18} uses the Bi-directional Attention Flow (BiDAF) \cite{DBLP:conf/iclr/SeoKFH17} model augmenting the bi-directional attention flow along with contextualized embeddings and self-attention. The modeling performs reasoning via a multi-layered bidirectional attention flow layer followed by a multi-layered biLSTM to identify the correct answer span. 
SDNet \cite{DBLP:journals/corr/abs-1812-03593} utilizes two bidirectional Recurrent Neural Networks (RNNs) \cite{rumelhart1986learning} to apply both self-attention and inter-attention between different layers in order to form the contextualized understanding of question and context. 

\subsubsection{Pre-trained Language Models}
Large-scale pre-trained language models such as BERT \cite{DBLP:conf/naacl/DevlinCLT19}, RoBERTa \cite{liu2019roberta}, and GPT \cite{radford2018improving} have become popular to achieve the state-of-the-art results on NLP tasks. While GPT is known for its language generation capabilities, BERT is famous for language understanding and has provided great results in machine comprehension tasks. 
One of the advantages of employing pre-trained language models is their capability to fuse both encoding and reasoning modules together. This results in a ready-to-tune architecture that hides the complex interactional nature between the given context and current question. However, incorporating previous context is a challenging task in pre-trained language models (particularly BERT) as it allows for only two segments in the input and the length of sequence is limited to 512. The more turns we try to append, the more context paragraph or history turns need to be truncated to be able to adapt to the model. The accurate modeling of the history results in better reasoning over the context.
The history integration challenge can be addressed using the following approaches:
\begin{itemize}
    \item Highlighting conversational history by embedding history answer embeddings in the contextual tokens as suggested in BERT-HAE \cite{DBLP:conf/cikm/QuYQZCCI19}. The embeddings are only added for those tokens that are present in the previous conversational history.

\item Using separate models for all 
the history turns to attend to the interaction between each turn and the given context as suggested by Ohsugi et al. \cite{ohsugi-etal-2019-simple}. The contextualized embeddings are then merged together to form an aggregated history-aware embeddings. These aggregated embeddings are then passed from  BiGRU to capture an inter-turn interaction before any prediction can be made. 

\item Introducing a reinforced backtracker in the model to filter out the unnecessary or irrelevant history turns instead of evaluating them as a whole as proposed by Qiu et al. \cite{qiu2021reinforced}. The selected turns along with the given passage forms an input to be provided to the BERT model.
\end{itemize}

Once the history turns have been integrated, BERT-based models calculate the probability of each word being the start word by generating a dot product between the final embedding and the start vector, followed by the application of softmax over all the words \cite{DBLP:conf/sigir/Qu0QCZI19}. Finally, the word with the highest probability value is selected. A similar process is employed to locate the final word in the given context. In \cite{qiu2021reinforced}, the model after predicting the answer span generates a reward to evaluate the utility of the history selection for answer prediction process. The computed reward, in turn, is utilized to update the policy network to maximize the accuracy of the model for the next cycle of prediction.

\subsubsection{Flow-based Models}
Another recent trend that has caught attention is the use of flow-based approaches in machine comprehension. A well-designed CMRC model should be able to grasp the flow of the conversation, i.e, knowing what topic is under discussion as well as facts and events relevant to it. Thus, the flow of conversation can be considered 
as a sequence of latent representations generated based on the token of source passage. These latent representations, generated during the reasoning of previous conversations, aid in the contextual reasoning of the current question. The main models based on flow architecture are described below.

FlowQA \cite{DBLP:conf/iclr/HuangCY19} utilizes the contextualized embeddings as the latent representations, a process often referred to as Integration Flow (IF). The process involves the sequential processing of the context tokens in parallel to the question turns (referred to as context integration) along with processing question turns sequentially parallel to context tokens (flow). The model utilizes multiple flow layers interweaved with attention first on the context and then on the question itself to come up with the reasoning for answer span.
FlowDelta \cite{DBLP:conf/acl-mrqa/YehC19} was introduced as an improved version in the flow series that utilizes the same architecture as FlowQA but achieves better accuracy. Instead of using the intermediate or latent representations, the model passes the information gain through the reasoning process. The information gain is nothing but the difference between the latent representations of the previous two layers. By modeling such difference, the model would better focus on the information hints present in the context.

The previously discussed flow approaches follow the concept of IF that does not really mimic a human's style of reasoning. The underlying reason is that they first perform reasoning in parallel for each question and then refine and enhance the reasoning across different turns. Graph Flow \cite{DBLP:conf/ijcai/0022WZ20}, on the other hand, constructs a dynamic context graph encoding not only the passage itself but also the question as well as the conversation
history. The model processes the flow by applying GNN on all the sequences of context graphs and the output is utilized when processing the next graph. To capture the contextual relationship between the words, a biLSTM is applied before providing the words as an input to GNN. The Graph Flow architecture alternates this mechanism with co-attention over the question and the GNN output.

\subsubsection{Open-retrieval Based Models}
Another recently introduced trend in the field of CMRC is the use of open-retrieval methods. The methods discussed above relies heavily on the given passage to extract or generate an answer. However, this seems impractical in real-world scenario since the availability of gold passage is not always possible. Thus, the model should be able to retrieve the relevant passages from a collection. The main models employing the open-retrieval architecture are discussed below:

Open-retrieval CQA (ConvQA) \cite{qu2020open} is first in the series of open retrieval models for CMRC. It consists of three main modules: i) a passage retriever, ii) a passage reranker, and iii) a passage reader. The three modules are based on Transformers \cite{NIPS2017_3f5ee243}. The passage retriever first extracts the top-K relevant paragraphs from a collection provided a current question and the previous history. The retriever is based on dual-encoder architecture that utilizes two separate ALBERT \cite{DBLP:conf/iclr/LanCGGSS20} encoders for passages and questions. The reranker and reader uses the same BERT encoder. The encoder transforms the input sequence consisting of question, history, and relevant passages into the contextualized representations to be utilized by reranker and reader for answer extraction. The reranker module conducts a list-wise reranking of the retrieved passages which serves as a supervision signal to fine-tune the encoder. In the end, answer span is predicted by the reader module by computing the probability of the tokens being a start/end token.

In ORConvQA, the model focuses on identifying and extracting short span-based answers. In information-seeking dialog, however, answers are relatively free-form and long which are difficult to extract. Weakly-supervised open-retrieval CQA (WS-ORConvQA) \cite{DBLP:conf/ecir/QuYCCKI21} is an extension of ORConvQA and introduces a learned weak supervision approach that can find and extract both span-based and free-form answers. And if the exact match is not found, the model tries to find a span in the retrieved passages that has the maximum overlap with the gold answer. Given a question and its conversation history, the passage retriever first extracts the relevant paragraphs from a collection. The retriever assigns a score based on the dot product of the representations of the questions and the passage. The reader then reads the top passages and produces an answer. The model works on weakly-supervised training approach. Given one of the retrieved passages and gold answer, the weak supervisor predicts a span in the passage as weak answer to provide weak supervision signals for training the reader. The reader is based on standard BERT-based machine comprehension model \cite{DBLP:conf/naacl/DevlinCLT19} that calculates the probability of tokens being a start and an end token. The final answer is selected by computing the sum of its retriever score and reader score. 
\afterpage{

\begin{landscape}
\centering
\scriptsize

\RaggedLeft
\begin{longtable}{|p{1.5cm}|p{2.5cm}|p{2.5cm}|p{3.8cm}|p{3.8cm}|p{2.3cm}|}

\caption{Recent studies on conversational machine reading comprehension (2016-2021).}
\label{tab:cmrct}
\\

\hline   \multicolumn{1}{|c|}{\textbf{Ref.}}  &  \textbf{History Selection} &  \multicolumn{1}{|c|}{\textbf{Encoder}}  & \multicolumn{1}{|c|} {\textbf  {History Modeling}} & \multicolumn{1}{|c|}{\textbf{Reasoning}}  &  \multicolumn{1}{|c|}{\textbf{Output Prediction}} \\ \hline 
\endfirsthead

\multicolumn{6}{c}%
{{\bfseries \tablename\ \thetable{} -- continued from previous page}} \\
\hline 
  \multicolumn{1}{|c|}{\textbf{Ref.}}  &  \textbf{History Selection} &  \multicolumn{1}{|c|}{\textbf{Encoder}}  & \multicolumn{1}{|c|}{ \textbf {History Modeling}} & \multicolumn{1}{|c|}{\textbf{Reasoning}}  &  \multicolumn{1}{|c|}{\textbf{Output Prediction}} \\ \hline 
\endhead

\hline \multicolumn{6}{|r|}{{Continued on next page}} \\ \hline
\endfoot

\hline 
\endlastfoot

BIDAF++ w/k-ctx \color{blue}$\textbf{\cite{DBLP:conf/emnlp/ChoiHIYYCLZ18}}$ & 
    \textit{k} history turns.
        &  \begin{tableitems}[nosep,after=\strut]
        \item GloVE for word embeddings.
        \item BiDirectional LSTM for contextual embeddings.
        \end{tableitems}
        & \begin{tableitems}[nosep,after=\strut]
                    
                         \item Encodes context tokens with history answer markers before passing on for reasoning.
                         \item Encode dialog turn number within the question embeddings. 
        \end{tableitems}  &Performs reasoning via multi-layered bidirectional attention flow layer followed by multi-layered biLSTM.& Span prediction.\\ \hline
        DrQA +PGNet \color{blue}$\textbf{\cite{DBLP:journals/tacl/ReddyCM19}}$ & 
      \textit {k} history turns.
        & Bidirectional LSTM
        & \begin{tableitems}[nosep,after=\strut]
                    
                         \item Appends the selected history turns to the source passage and current question.
        \end{tableitems}  &DrQA model first point towards the evidence in the given text, PGNet then transform the evidence into the answer. & Free-form answers\\ \hline
        SDNet \color{blue}$\textbf{\cite{DBLP:journals/corr/abs-1812-03593}}$ & 
     \textit{k} history turns
        & \begin{tableitems}[nosep,after=\strut]
                    
                         \item Word embeddings using GloVe.
                         \item Contextualized embeddings using BERT.
        \end{tableitems}
        & \begin{tableitems}[nosep,after=\strut]
                    
                         \item Appends the selected history turns to the source passage and current question.
        \end{tableitems}  &Utilizes both self-attention and inter attention in multiple layers using biDirectional LSTM to reason across the given context.& Span prediction\\ \hline
        \hline
        BERT-HAE \color{blue}$\textbf{\cite{DBLP:conf/sigir/Qu0QCZI19}}$ & 
      \textit{k} history turns but found optimal answer in 5 and 6 history turns
        & BERT-generated embeddings
        & \begin{tableitems}[nosep,after=\strut]
                    
                         \item Introduce history answer marker layer to the context token is present in any conversational history answer or not.
        \end{tableitems}  & \begin{tableitems}[nosep,after=\strut]
                    
                         \item BERT generates a representation for each token based on the embeddings for position, segment, and tokens.
                         \item The model then computes the probability of tokens in a given paragraph of being a start and end token of the answer span.
        \end{tableitems} & Span prediction\\ \hline
        BERT-HAM \color{blue}$\textbf{\cite{DBLP:conf/cikm/QuYQZCCI19}}$ & Dynamic history selection policy
        & Bert-based embeddings on both word and sequence level
        & \begin{tableitems}[nosep,after=\strut]
                        \item Encode context tokens with dialog-turn encoded variant of HAE called \textit{Positional-HAE}.
        \end{tableitems}  & History attention module assigns weight to each token level and sequence level representation. And then aggregated representations of both are obtained that are further used for answer prediction. & \begin{tableitems}[nosep,after=\strut]
                    
                         \item Span prediction.
                         \item Dialog-act prediction.
        \end{tableitems}\\ \hline
        BERT w/k-ctx \color{blue}$\textbf{\cite{ohsugi-etal-2019-simple}}$ & 
      \textit{k} history turns
        & Contextualized paragraph representations
independently conditioned with each question and
each answer generated using BERT.
        & \begin{tableitems}[nosep,after=\strut]
                    
                         \item Appends history QA pair to the current question with each QA pair conditioned on the source paragraph.
                         \item The model then concatenates the resulting sequences to form a uniform representation.
        \end{tableitems}  &The concatenated result is then passed through the BiGRU for span prediction. & \begin{tableitems}[nosep,after=\strut]
                    
                         \item Span prediction
                         \item Answer type prediction (Yes, no, unanswerable).
        \end{tableitems}  \\
        \hline
        Env-ConvQA \color{blue}$\textbf{\cite{qiu2021reinforced}}$ & 
     dynamic \textit{k} history turns
        & BERT-generated embeddings.
        & \begin{tableitems}[nosep,after=\strut]
                    
                         \item Prepends selected subset of history QA pair and passage to the current question.
                         \item The model then concatenates the resulting sequences to form a uniform representation.
        \end{tableitems}  & \begin{tableitems}[nosep,after=\strut]
                    
                         \item BERT generates a representation for each token based on the embeddings for position, segment, and tokens.
                         \item The model then computes the probability of tokens in a given paragraph of being a start and end token of the answer span.
                         \item After answer prediction, the model generates a reward to evaluate the role of selected history turns and update the policy network accordingly.
        \end{tableitems} & 
                    
                          Span prediction \\

        \hline \hline
        FlowQA \color{blue}$\textbf{\cite{DBLP:conf/iclr/HuangCY19}}$ & \textit{k} history turns.
        & Uses ELMo to generate contextual embeddings before passing it to IF layer. 
        &  Integrates both QA pairs and  the intermediate context representation from conversation history called \textbf{FLOW.}
         & Employ multiple integration flow layers with alternating cross and self-attention to perform reasoning. & Span prediction\\ \hline
         Graph Flow \color{blue}$\textbf{\cite{DBLP:conf/ijcai/0022WZ20}}$ & 
      prepends \textit{N} question answer pairs to the current question.
        & GloVE and 1024-dim BERT embeddings
        & Encodes history QA pairs into contextual graphs.
          &BiLSTM is utilized for the context integration and the GNNs are used to capture the contextual interaction. & Span prediction\\ \hline
          FlowDelta \color{blue}$\textbf{\cite{DBLP:conf/acl-mrqa/YehC19}}$ & 
      \textit{k} history turns
        & Uses ELMo to generate contextual embeddings before passing it to IF layer.
        & Integrates both QA pairs and  the intermediate context representation from conversation history called \textbf{FLOW.}
                  & Model passes the information gain (the difference between the latent representations of last two layers) to let the model focus more precisely on the context.  & Span prediction\\
                  \hline
                  \hline
                  ORConvQA \color{blue}$\textbf{\cite{qu2020open}}$ & 
      \textit{k} history turns
        & Uses ALBERT to generate contextual embeddings before passing it to reader and reranker modules.
        & \begin{tableitems}[nosep,after=\strut]
                    
                         \item Appends history questions to the current question.
                         \item The model uses two encoders, one for encoding current question with its history and other for encoding relevant passages.
        \end{tableitems}
                  & \begin{tableitems}[nosep,after=\strut]
                    \item Employs fully-supervised setting for the training of the reader.
                         \item The top-retrieved passages are then fed to the reranker and reader for a concurrent learning of all model components.
                         \item The reader predicts an answer by computing scores of each token being the start token and the end token.
        \end{tableitems}  & Span prediction\\
        \hline
                  
                  WS-ORConvQA \color{blue}$\textbf{\cite{DBLP:conf/ecir/QuYCCKI21}}$ & 
      \textit{k} history turns
        & Uses ALBERT to generate contextual embeddings before passing it to reader module.
        & \begin{tableitems}[nosep,after=\strut]

                         \item Appends history questions to the current question.
                         \item The model uses two encoders, one for encoding current question with its history and other for encoding relevant passages.
                         \item The retriever generates a score based on the dot product of the representations of the question and the passage.
        \end{tableitems}
                  & \begin{tableitems}[nosep,after=\strut]
                    \item Employs weakly-supervised setting for the training of the reader.
                         \item The top-retrieved passages are then fed to the reader.
                         \item The reader computes the probabilities of the true start and end tokens among all the tokens from the top passages.
                         \item The answer span is selected on the basis of sum of retriever score and reader score.
        \end{tableitems}  & \begin{tableitems}[nosep,after=\strut]
                    
                         \item Span prediction.
                         \item Free-form answer.
        \end{tableitems}\\

\hline 
\end{longtable}
\vfill
\end{landscape}
}

\subsection{Output Prediction}
The common trends that have been observed for the answer prediction module include span prediction, free-form answer prediction, and dialog acts prediction. For span prediction, the probabilities of tokens being the end and start token is calculated. For unanswerable questions, a token, UNANSWERED, is appended at the end of each passage in QuAC. The model learns to predict this token if it finds the question unanswerable. A sequence-level aggregated representation is used to calculate dialog-act prediction and the modeling of history dialog-acts is not required for the prediction of this task.

The categorization of the architecture based on the techniques used in each module is 
summarized in Table~\ref{tab:cmrct}.

\section{Datasets for Conversational Question Answering}
\label{dataset}
One driver for the rapid growth in the field of CQA is the emergence of large-scale conversational datasets for both knowledge-base and machine comprehension. 
Constructing a high-quality dataset is equally significant as optimizing CQA-based architectures. 
In this section, we collect and compare the major datasets in the area of CQA. 
\subsection{Datasets for Sequential KB-QA}
Most of the datasets for sequential KB-QA deals with simple questions, wherein each of them can be answered using a single tuple in the knowledge graph. However, in practice, 
a system can encounter a more complicated form of questions requiring it to use logical and comparative reasoning to come up with an accurate answer. The point worth noting is that unlike the simple questions, the complicated questions require access to the larger subgraph of the KG. For example, to answer the question, ``\textit{Which country has the highest peak, Nepal or India?}'', one needs to find i) the highest peak in Nepal, ii) the highest peak in India, and finally, iii) the comparison of both the peaks to come up with the right answer. 

Similar to the field of CMRC, sequential KB-QA saw its rise after the introduction of sequential QA datasets namely  \textit{Sequential Question Answering (SQA)} \cite{DBLP:conf/acl/IyyerYC17}, \textit{Complex Sequential Question Answering (CSQA)} \cite{DBLP:conf/aaai/SahaPKSC18}, and \textit{ConvQuestions} \cite{DBLP:conf/cikm/ChristmannRASW19}. These datasets have facilitated the process of answering complex questions, thus 
supporting 
a number of researches. A high-level comparison based on their common characteristics is presented in Table~\ref{tab:characteristicsSQA}.

\subsubsection{SQA}
The main idea behind the creation of SQA is to decompose the complex questions and convert them into a series of inter-linked sequential questions to give a touch of natural conversation. 

\paragraph{Dataset Collection:}
As described in \cite{DBLP:conf/acl/IyyerYC17}, 
the SQA dataset has been collected via crowdsourcing by leveraging WikiTable Questions (WTQ)\footnote{https://github.com/ppasupat/WikiTableQuestions}, which contains highly compositional questions associated with HTML tables from Wikipedia. Each crowdsourcing task contains a long and complex question originally from WTQ as the question intent. The workers are asked to compose a sequence of simpler but inter-related questions that lead to the final intent. The answers to the simple questions are subsets of the cells in the table.

\paragraph{Dataset Analysis:}
SQA consists of 6,066 unique question sequences containing 17,553 
question-answer pairs resulting in an average of 2.9 questions per sequence. The questions are identified into three different classes: i) \textit{column selection} questions, wherein the answer is the entire column of the table and constitutes 23\% of the questions in SQA, ii) \textit{subset selection} questions where the answer is the subset of the previous question's answer and contributes 27\% of the questions in the  dataset, and iii) \textit{row selection} questions where answers to the questions appear in the same rows but in different columns, making 19\% of the dataset. 

\paragraph{Evaluation:}
For the system to be evaluated, the overall accuracy, sequence accuracy (the percentage of sequences
for which every question is answered correctly), and positional accuracy (accuracy at each position in a sequence) are calculated.
With that said, all systems struggle to
 correctly answer all questions within a sequence,
despite the fact that each question is
simpler on average than those in WTQ.

\begin{scriptsize}
\RaggedLeft
	\begin{table}[tbp!]
		\caption{A comparison of the sequential KB-QA datasets SQA\cite{DBLP:conf/acl/IyyerYC17}, CSQA\cite{DBLP:conf/aaai/SahaPKSC18}, and ConvQuestions\cite{DBLP:conf/cikm/ChristmannRASW19} based on different characteristics as defined in their respective papers.}
		\label{tab:characteristicsSQA}
	\begin{tabular}{|p{2.5cm}|p{2.5cm}|p{2.5cm}|p{3.0cm}|}
		\hline
		\multicolumn{1}{|c|}{\textbf{Characteristics}}  & \multicolumn{1}{|c|}{\textbf{SQA}}  &  \multicolumn{1}{|c|}{\textbf{CSQA}} &
		\multicolumn{1}{|c|}{\textbf{ConvQuestions}}\\
		\hline 
	 \textbf{Data Source}& WikiTableQuestions & WikiData  & WikiData (consisting of 5 domains i.e. books, movies, soccer, music, and tv series) \\
\hline
 \textbf{Conversational Setup} &Three workers who were asked to decompose complex sentences into a sequence of simpler sequential sentences & Pair of in-house annotators where annotator acts as a \textit{user} and the other as a \textit{system} to provide answers or ask clarification questions  & Master workers from AMT paired together where they were asked to provide answers vis web search \\
 \hline
      \textbf{Nature of QAs} & Simple  & Complex inter-related as well as simple & Complex \\
      \hline
       
      \textbf{Question Types} & Factoid  & Factoid & Factoid and non-opinionated \\
\hline
     \textbf{Requires Reasoning?} & No & Yes & Yes\\ \hline
     \textbf{Max turns per dialog} & N/A & 8.5 & 5\\ \hline
     
      \textbf{Total Number of Questions} & 15, 553 & 1.6 M & N/A \\ \hline
      \textbf{Total Number of Dialogs} & 6,066 & 200K & 11,200\\ 
     
      \hline
		
	\end{tabular}
\end{table}
\end{scriptsize}
\subsubsection{CSQA}
The CSQA dataset \cite{DBLP:conf/aaai/SahaPKSC18} consists of 200K QA dialogs for the task of complex sequential question answering.
CSQA combines two sub-tasks: i) answering factoid questions through complex reasoning over a large-scale KB, and ii) learning to converse through a sequence of coherent QA pairs. CSQA calls for a sequential KB-QA agent that combines many technologies including i) parsing complex natural language queries, ii) using conversation context to resolve co-reference and ellipsis in user utterances like the belief tracker, iii) asking for clarification questions for ambiguous queries, like the dialog manager, and iv) retrieving relevant paths in the KB to answer questions.

\paragraph{Dataset Collection:}
Each dialog is prepared in a two-in-house-annotators setting, one being a \textit{user} and the other acting as a \textit{system}. A user's role is to ask questions and a system's job is to answer the questions or asks for clarification if required. The idea is to establish the understanding of the simple and complex questions that can be asked by the annotators over a knowledge graph. These could then be abstracted to templates and utilized to instantiate more queries involving different objects, subjects, and relations. Apart from asking and answering simple questions (that requires only a single tuple to generate an answer), the annotators 
come 
up with questions involving logical and comparative operators like AND, OR, NOT, ==, and $>$=, 
resulting in more complex questions to judge model's performance. The examples of such questions are ``\textit{Which country has more population than India?}'', and ``\textit{Which cities of India and Pakistan have River Indus passing through them?}''.
After collecting both simple and complex questions, the next step is to create coherent conversations involving these QA pairs. The resulting conversation should have i) linked subsequent QA pairs, and ii) the conversation should contain the necessary elements of a conversation such as confirmation, clarification, and co-references. 

\paragraph{Dataset Analysis:}
The dataset consists of 200K dialogs and a total of 1.6 million turns. On average, the length of a user's questions is 9.7 words and a system's response is based on 4.74 words.

\paragraph{Evaluation:}
Different evaluation metrics are used to evaluate the different question types. For example to measure the accuracy of simple questions (consisting of indirect questions, co-references, ellipsis), logical reasoning, and comparative reasoning, both precision and recall are used. When dealing with quantitative reasoning and verification (boolean) questions, F1 score is utilized. For clarification questions, BLEU-4 score is used.

\subsubsection{ConvQuestions}
ConvQuestions \cite{DBLP:conf/cikm/ChristmannRASW19} has been published recently to further aid the field of sequential KB-QA. It consists of 11,200 distinct conversations from five different domains, i.e, books, movies, soccer, music, and TV-series. The questions are asked with minimal syntactic guidelines to maintain the natural factor of the questions. The questions in ConvQuestions are sourced from WikiData and the answers are provided via Web search. The questions in the dataset pose different challenges that need to be addressed including incomplete cues, anaphora, indirection, temporal reasoning, comparison, and existential.

\paragraph{Dataset Collection:}
Each dialog is prepared as a conversation generation task by the workers of AMT wherein they were asked to base their conversation on the five \textit{sequential} questions from any 
domain 
of their choice. 
To make sure that the conversations are carried out as naturally as possible, the Turkers were asked not to interleave the questions and neither permute the order of follow-up questions to generate a large volume. Furthermore, the paraphrases of the questions were also collected to provide two versions of the questions. This would allow the data to be augmented with several interesting variations which, in turn, improves the robustness of the system. To make the dataset more closely related to real-world challenges, participants were encouraged to ask the complex questions.

\paragraph{Dataset Analysis:}
The dataset consists of 11,200 conversations each comprising of 5 turns. The average length of the first and follow-up questions were 9.07 and 6.20 words, respectively. Question entities and expected answers have a balanced distribution among non-human types (books, stadiums, TV-series) and humans (actors, artists, authors). Context expansion is the key for finding out the correct answer in ConvQuestions as the average KG distance from the original seed to the answer is 2.30. The question type consists of characteristics such as comparisons, temporal reasoning, and anaphora, 
to make it more closely related to real-world challenges. 

\paragraph{Evaluation:}
Since each question in the dataset has exactly one or at most three correct answers, it uses standard metrics of Precision at the top rank (P@1). The other metrics include Mean Reciprocal Rank (MRR) and Hit@5. Hit@5 measures the fraction of times a correct answer is identified within the top-5 positions.

\subsection{Datasets for Conversational Machine Comprehension}
Generally, the datasets for machine reading comprehension falls into three categories based on the type of answer they provide:

\begin{itemize}
\item \textit{Multiple-choice option} datasets provide text-based multiple choice question and expect the model to identify the right answer out of the available options. The examples of such datasets include RACE \cite{DBLP:conf/emnlp/LaiXLYH17}, MCTest \cite{DBLP:conf/emnlp/RichardsonBR13}, and MCSript \cite{DBLP:conf/lrec/0002MRTP18}, 

\item \textit{Descriptive answer} datasets allow answers to be in any free-form text. Such datasets are useful in situations, wherein the questions are implicit and may require the use of common sense or world knowledge. The examples include MS Marco \cite{DBLP:conf/nips/NguyenRSGTMD16} and Narrative QA \cite{DBLP:journals/tacl/KociskySBDHMG18}, and 
\item \textit{Span prediction} or \textit{extractive} datasets 
require the model to extract the correct answer span from the given source passage. Such datasets provide better natural language understandability and easy evaluation of the task. SQuAD \cite{DBLP:conf/emnlp/RajpurkarZLL16}, TriviaQA \cite{DBLP:conf/acl/JoshiCWZ17}, and NewsQA \cite{DBLP:conf/rep4nlp/TrischlerWYHSBS17} are some of the popular examples of extractive datasets.
\end{itemize}

 CoQA \cite{DBLP:journals/tacl/ReddyCM19} and QuAC \cite{DBLP:conf/emnlp/ChoiHIYYCLZ18}, the two datasets for CMRC, comes under the category of span-prediction datasets. Apart from these two datasets, there is another CMRC dataset, ShARC \cite{DBLP:conf/emnlp/SaeidiBL0RSB018}, which requires the understanding of a rule-text to answer a few inter-linked and co-referenced questions. These generated questions need to be answered using reasoning on the basis of background knowledge. However, this dataset does not really follow the definition of CMRC and is hence ignored. A summarized comparison pertaining to significant characteristics of both CoQA and QuAC is presented in Table~\ref{cmrc}. 

\begin{scriptsize}
\RaggedLeft
	\begin{table}[tbp!]
		\caption{A comparison of the multi-turn conversational
datasets-CoQA \cite{DBLP:journals/tacl/ReddyCM19} and QuAC \cite{DBLP:conf/emnlp/ChoiHIYYCLZ18} based on different characteristics as defined in their respective papers.}
		\label{cmrc}
	\begin{tabular}{|p{3.5cm}|p{4.0cm}|p{3cm}|}
		\hline
		\multicolumn{1}{|c|}{\textbf{Characteristics}}  & \multicolumn{1}{|c|}{\textbf{CoQA}}  &  \multicolumn{1}{|c|}{\textbf{QuAC}} \\
		\hline 
		\textbf{Data Source} &  Passages collected
from 7 diverse domains e.g. children
stories from MCTest,
news articles from
CNN, Wikipedia articles etc. & Sections from Wikipedia articles filtered in the “people” category associated with subcategories like culture, animal, geography, etc. \\ \hline
\textbf{Conversational Setup }  & Questioner-answerer setting where both have access to the entire context. & Teacher-Student setting where the teacher has access to the full context for answering, while the student has only the title and summary of the article\\ \hline
\textbf{Requires External Knowledge?} & Yes & No\\
		\hline
		\textbf{Question Type} & Factoid & Open-ended, highly contextual\\ \hline
		\textbf{Answer Type} & Free-form with an extractive rationale. & Extractive span which can be
yes/no or ‘No Answer’.\\ \hline
		\textbf{Dialog Acts} & No & Yes\\ \hline
		\textbf{Max Turns per Dialog }& 15 & 11 \\ \hline
		\textbf{Unanswerable Questions} & Yes & Yes \\ \hline
		\textbf{Total Number of Questions} & 126K & 100K \\ \hline
		\textbf{Total Number of Dialogs} & 8K & 14K \\ \hline
		
	\end{tabular}
\end{table}
\end{scriptsize}

\subsubsection{CoQA}
CoQA was introduced by Reddy et al. \cite{DBLP:journals/tacl/ReddyCM19} to measure a machine's ability to participate in a QA style conversation. The dataset was developed with three objectives in mind. The first is the nature of questions in human conversations. In this dataset, every question except the first one is dependent on the conversation history to make it more similar to the real-life setting of human conversation.
The second goal of CoQA is to maintain the naturalness of answers in a conversation. Many existing datasets limit answers to be found in the given source passage. However, such a setting does not always ensure natural answers. In CoQA, the authors address this issue by proposing free-form answers while providing a text-span from the given passage as a rationale to the answer. 

The third goal of CoQA is to facilitate the development of CQA systems across multiple domains. The existing QA datasets mainly focus on a single domain which results in complications to test the generalization capabilities of the existing systems. Thus, CoQA extends its domains, i.e., each with its own data source. These domains include articles based on literature extracted from Project Gutenberg\footnote{https://www.gutenberg.org/}, children's stories taken from MCTest \cite{DBLP:conf/emnlp/RichardsonBR13}, Wikipedia articles\footnote{https://www.wikipedia.org/}, Reddit articles from Writing Prompt \cite{DBLP:conf/acl/LewisDF18}, middle and high school English exams taken from \cite{DBLP:conf/emnlp/LaiXLYH17}, science articles derived from Ai2 science question \cite{DBLP:conf/aclnut/WelblLG17}, and news articles taken from CNN \cite{DBLP:conf/nips/HermannKGEKSB15}. Evaluation and Reddit are used for out-of-domain evaluation only.

\paragraph{Data Collection:}
Each conversation is prepared in a two annotator setting, i.e., one being a questioner and the other being an answerer. The platform of Amazon Mechanical Turk (AMT)\footnote{https://www.mturk.com/} is used to pair workers on a passage through the ParlAI MTurk API \cite{DBLP:conf/emnlp/MillerFBBFLPW17} and both the annotators have full access to the passage. 

\paragraph{Dataset Analysis:}
The dataset consists of 127K conversation turns gathered from 8K conversations over text passages. The average length of a conversation is 15 turns and each turn consists of a question and an answer. The distribution of CoQA is spread across multiple question types. Prefixes like \textit{did}, \textit{where}, \textit{was}, \textit{is}, and \textit{does}
are very frequent in the dataset. Also, almost every sector of CoQA contains co-references which shows that it is highly conversational. What makes conversations in CoQA even more human-like is that sometimes they just feature one-word questions like “who?” or “where?” or even “why?”. This shows that questions are context-dependant, and in order to answer correctly, the system needs to go through the previous history turns to understand the question.

\paragraph{Evaluation:}
The main evaluation metric for the dataset is macro-average F1 score of word overlap and is computed separately for in-domain and out-of-domain as well.

\subsubsection{QuAC}
In an information-seeking dialog, the students keep 
asking their teacher questions 
for clarification about a particular topic. 
This idea forms the basis for this newly introduced dataset,  Question Answering in Context (QuAC). Modeling such inter-related questions can be complex as the questions can be elliptical, highly context-dependent, and even sometimes unanswerable. To promote learning in such a challenging situation, QuAC presents a rich set of 14K crowd-sourced QA dialogs (consisting of 100K QA pairs). 

\paragraph{Dataset Collection:}
 The nature of interaction in QuAC is of student-teacher where the teacher has the access to the source paragraph. A student only provided with the heading of the paragraph aims to gain as much knowledge about its content as possible by asking multiple questions. The teacher tries to answer the questions by extracting correct answer spans from the source passage. Also, the teacher uses dialog acts as feedback to the students (i.e., may or may not ask a follow-up question) which results in more productive dialogs. 

\paragraph{Dataset Analysis:}
The dataset has long answers of maximum of 15 tokens which is an improvement over SQuAD and CoQA. Another factor worth noting is that frequent question types in QuAC are based on \textit{Wh} words which makes the questions more open-ended, in contrast to the other QA datasets where questions are more factoid. Furthermore, 86\% of the questions are highly contextual, i.e., they require the model to re-read the context to resolve the co-references. Out of these questions, 44\% refer to entities or events in the dialog history whereas 61\% refer to the subject of the article.   

\paragraph{Evaluation:}
Besides evaluating the accuracy using F1 score, QuAC also utilizes human equivalence score (HEQ) 
to measure a system's performance by finding the percentage of exceeding or matching an average human's performance. HEQ-Q and
HEQ-D are, therefore, HEQ scores with the instances as
questions and dialogs respectively.

\section{Research Trends and Open Challenges}
\label{challenges}
CQA is a rapidly evolving field. This paper surveys 
neural approaches 
that 
have been recently introduced to cater to the challenges pertaining to CQA. 
These CQA systems have the potential to be successfully utilized in 
practical applications: 
\begin{itemize}
    \item The KB-QA based systems allow users to access a series of information via conversation without even composing complex SQL queries. From commercial perspective, these KB-QA based systems can be employed either in open-domain QA (pertaining to worldly knowledge) or in closed-domain QA (such as in the medical field). A user 
    does not 
    have to access multiple sources of information, rather one agent would suffice 
    her 
    all information needs.
    
    \item CQA systems provide simplified conversational search (ConvSearch) setting \cite{DBLP:conf/sigir/Qu0QCZI19} which has the strongest potential to become more popular than the traditional search engines such as Google or Bing, which unlike a user's expectations of getting a concise answer, provides a list of probable answers/solutions. These conversational systems can potentially be used for learning about a topic, planning an activity, seeking advice or guidance, and making a decision. 
    
    \item The conversational agents play a significant role in facilitating smooth interaction with users. One of the conceivable applications
    could be customer support systems where 
    a 
    user does not have to go through the entire website and looks for the desired information. 
\end{itemize}

As an emerging research area with many significant promising applications, CQA techniques are still not mature yet with many open issues remaining. In this section, we discuss several prominent ones: 

\begin{itemize}
    \item The role of context to be selected plays a significant role in providing accurate answers in CQA. With richer conversational scenarios, a number of contextual features need to be considered including personal context, social context, and task context. General research questions regarding contextual information in CQA include: ``\textit{What are the effective strategies and models to collect and integrate contextual information?}'', ``\textit{Are knowledge graphs sufficient enough to capture and represent this information?}'', and ``\textit{Do we need to incorporate the entire context or a relevant chunk would be enough to find the correct answer?}''.
    
    Different models attempt to incorporate context in different ways. Out of all the history selection methods, the dynamic history selection mechanism proposed by Qu et al. \cite{DBLP:conf/cikm/QuYQZCCI19} is more compelling and intuitive.
    As far as 
    the flow-methods are concerned, they consider the latent representations of the entire context to deal with the varying conversational aspect. Similarly, for sequential KB-QA, the authors in \cite{DBLP:conf/nips/GuoTDZY18} proposed the use of dialog manager to collect and maintain the previous utterances. 
    
    \item Information-seeking behaviors need to be modeled for CQA setting as it provides users with the opportunity to obtain more information about the topics of their interests. The research questions related to information-seeking behavior that needs to be explored include: ``\textit{What optimal structure for clarification questions can be used to better understand the users' information need?}'' and ``\textit{What effective strategies can be employed to design such clarification questions?}''.
    
    \item Interpretability of a question plays a significant role when finding an answer 
    for it. 
    In the existing CQA systems, the models are anticipated to provide the answers to the questions without having to explain as to why and how they deduced an answer, making it difficult to understand the source and reason of an answer. CoQA is the only CQA dataset that provides reasoning for the provided answer. Another model, Cos-E \cite{DBLP:conf/acl/RajaniMXS19}, generates commonsense reasoning explanations for the deduced answer. Regardless of the fact whether or not the complete interpretability of CQA models is required, we can safely say that an understanding of the working of the internal model up to a certain extent can greatly help and improve the design of neural network systems in the future. 
    
    \item Commonsense reasoning is a long-standing challenge in Conversational AI, i.e., whether it is incorporating the commonsense in dialog systems or QA systems. Commonsense reasoning refers to the ability of an individual to make day-to-day inferences by using or assuming basic knowledge about the real-world. However, the CQA systems proposed so far work on pragmatic reasoning, i.e., finding the intended meaning(s) from the provided context because commonsense knowledge is often not explicitly explained in the data sources (i.e., KB-QA or CMRC dataset). Despite single-turn QA systems almost achieving human-level performance, the implementation of commonsense reasoning is still not very common. There are only a few research works that take commonsense reasoning into consideration when performing single-turn QA \cite{DBLP:conf/lrec/0002MRTP18, DBLP:conf/emnlp/HuangBBC19}.
    
  There has been an increasing trend to incorporate commonsense reasoning into the single-turn MRC over the past few years. But when it comes to utilizing commonsense reasoning in CMRC, no successful attempt has been made. This may probably be owing to the fact that commonsense reasoning requires questions that needs some prior knowledge or background which the current CMRC datasets do not provide. 
  When it comes to single-turn KB-QA, there are a number of prominent researches that utilize commonsense in a QA process \cite{DBLP:conf/emnlp/LinCCR19,DBLP:conf/aaai/SharmaG19, DBLP:conf/aaai/LvGXTDGSJCH20}. Another effort was done by CoMET \cite{bosselut-etal-2019-comet}, wherein a Transformer to generate commonsense knowledge graphs was employed. Knowledge graphs like ConceptNet \cite{speer2017conceptnet} and ATOMIC \cite{sap2019atomic} have been designed to facilitate the implementation of commonsense in KB-QA systems. The field of sequential KB-QA remains untouched primarily because of the reason that the majority of existing methods lack the absence of connections between 
  concepts \cite{DBLP:conf/nlpcc/ZhongTDZWY19}. 
  
  \item Lack of inference capability is one of the reasons why QA struggles with generating the correct answers. Most of the existing CQA systems are based on semantic relevance between question and the given context which limits a model's capability to reason. An example discussed by Liu et al. \cite{8651505} depicts that provided the context, ``\textit{five people on board and two people on the ground died}'', the system was not able to provide the correct answer ``\textit{seven}'' to the question ``\textit{how many people died?}''. Thus, how to design systems with strong inference ability is still an open issue and calls for further research. 
      
\end{itemize}
  
  \section{Conclusion}
  \label{conclusion}
  The Conversational Question Answering (CQA) systems have been 
  emerging as a main technology 
  to close the interactional gap between machines and humans owing to the advancements in pre-trained language modeling and the introduction of conversational datasets. This progress simplifies the development and progress of application areas such as online customer support, interactions with IoT devices in smart spaces, search engines, thus enabling CQA to realize its social and economic impacts. The effective incorporation of contextual information, the ability to infer the questions and ask efficient clarification questions are the main challenges pertaining to the field of CQA.
  
  Our survey on over 80 academic works 
  in the field of CQA, i.e., from 2016 to 2021, confirms the thriving expansion of this exciting field. In this survey, we have comprehensively discussed the field of CQA, which is further subdivided into i) sequential KB-QA, and ii) 
  Conversational Machine Reading Comprehension (CMRC). The general architecture of each of the category is decomposed into modules and prominent techniques employed in each module 
  have been  discussed. We subsequently introduced and discussed the representative datasets based on their characteristics. Finally, we discussed the potential applications of CQA and the identified future research directions that need to be explored for realizing natural conversations. 
%
We anticipate that this literature survey 
will serve as a quintessence for the researchers and pave a way forward for streamlining the research in this important area. 

\begin{acknowledgements}
Munazza Zaib sincerely acknowledges the generous support of the Macquarie University, Sydney, Australia for funding this research work via its International Macquarie University Research Excellence Scholarship (Allocation No. 20201589).
Wei Emma Zhang and Quan Z. Sheng have been partially supported by Australian Research Council (ARC) Discovery Grant DP200102298.
\end{acknowledgements}

%
%

\bibliographystyle{spbasic}
\bibliography{springer}   

\begin{thebibliography}{118}
\providecommand{\natexlab}[1]{#1}
\providecommand{\url}[1]{{#1}}
\providecommand{\urlprefix}{URL }
\expandafter\ifx\csname urlstyle\endcsname\relax
  \providecommand{\doi}[1]{DOI~\discretionary{}{}{}#1}\else
  \providecommand{\doi}{DOI~\discretionary{}{}{}\begingroup
  \urlstyle{rm}\Url}\fi
\providecommand{\eprint}[2][]{\url{#2}}

\bibitem[{Bahdanau et~al.(2015)Bahdanau, Cho, and
  Bengio}]{DBLP:journals/corr/BahdanauCB14}
Bahdanau D, Cho K, Bengio Y (2015) Neural machine translation by jointly
  learning to align and translate. In: Proceedings of the 3rd International
  Conference on Learning Representations, {ICLR} 2015, pp 01--15

\bibitem[{Bao et~al.(2016)Bao, Duan, Yan, Zhou, and
  Zhao}]{DBLP:conf/coling/BaoDYZZ16}
Bao J, Duan N, Yan Z, Zhou M, Zhao T (2016) Constraint-based question answering
  with knowledge graph. In: Proceedings of the 26th International Conference on
  Computational Linguistics, {COLING} 2016, pp 2503--2514

\bibitem[{Beaver et~al.(2020)Beaver, Freeman, and Mueen}]{beaver2020towards}
Beaver I, Freeman C, Mueen A (2020) Towards awareness of human relational
  strategies in virtual agents. In: Proceedings of the 34th Conference on
  Artificial Intelligence, {AAAI} 2020, pp 2602--2610

\bibitem[{Bhutani et~al.(2020)Bhutani, Zheng, Qian, Li, and
  Jagadish}]{bhutani-etal-2020-answering}
Bhutani N, Zheng X, Qian K, Li Y, Jagadish H (2020) Answering complex questions
  by combining information from curated and extracted knowledge bases. In:
  Proceedings of the First Workshop on Natural Language Interfaces, pp 1--10,
  \doi{10.18653/v1/2020.nli-1.1}

\bibitem[{Bollacker et~al.(2008)Bollacker, Evans, Paritosh, Sturge, and
  Taylor}]{bollacker2008freebase}
Bollacker K, Evans C, Paritosh P, Sturge T, Taylor J (2008) Freebase: a
  collaboratively created graph database for structuring human knowledge. In:
  Proceedings of the 2008 ACM SIGMOD International Conference on Management of
  Data, pp 1247--1250

\bibitem[{Bosselut et~al.(2019)Bosselut, Rashkin, Sap, Malaviya, Celikyilmaz,
  and Choi}]{bosselut-etal-2019-comet}
Bosselut A, Rashkin H, Sap M, Malaviya C, Celikyilmaz A, Choi Y (2019) {COMET}:
  commonsense transformers for automatic knowledge graph construction. In:
  Proceedings of the 57th Annual Meeting of the Association for Computational
  Linguistics, {ACL} 2019, pp 4762--4779, \doi{10.18653/v1/P19-1470}

\bibitem[{Bouziane et~al.(2015)Bouziane, Bouchiha, Doumi, and
  Malki}]{bouziane2015question}
Bouziane A, Bouchiha D, Doumi N, Malki M (2015) Question answering systems:
  survey and trends. Procedia Computer Science 73:366--375

\bibitem[{Budzianowski et~al.(2018)Budzianowski, Wen, Tseng, Casanueva, Stefan,
  Osman, and Ga{\v{s}}i\'c}]{budzianowski2018multiwoz}
Budzianowski P, Wen TH, Tseng BH, Casanueva I, Stefan U, Osman R, Ga{\v{s}}i\'c
  M (2018) {M}ulti{WOZ} - a large-scale multi-domain {W}izard-of-{O}z dataset
  for task-oriented dialogue modelling. In: Proceedings of the Conference on
  Empirical Methods in Natural Language Processing, {EMNLP} 2018, pp
  5016--5026, \doi{10.18653/v1/D18-1547}

\bibitem[{Cascante{-}Bonilla et~al.(2019)Cascante{-}Bonilla, Sitaraman, Luo,
  and Ordonez}]{DBLP:journals/corr/abs-1908-03180}
Cascante{-}Bonilla P, Sitaraman K, Luo M, Ordonez V (2019) Moviescope:
  Large-scale analysis of movies using multiple modalities. arXiv:190803180

\bibitem[{Chen et~al.(2017)Chen, Fisch, Weston, and
  Bordes}]{chen-etal-2017-reading}
Chen D, Fisch A, Weston J, Bordes A (2017) Reading {W}ikipedia to answer
  open-domain questions. In: Proceedings of the 55th Annual Meeting of the
  Association for Computational Linguistics, {ACL} 2017, pp 1870--1879,
  \doi{10.18653/v1/P17-1171}

\bibitem[{Chen et~al.(2020)Chen, Wu, and Zaki}]{DBLP:conf/ijcai/0022WZ20}
Chen Y, Wu L, Zaki MJ (2020) Graphflow: exploiting conversation flow with graph
  neural networks for conversational machine comprehension. In: Proceedings of
  the 29th International Joint Conference on Artificial Intelligence, {IJCAI}
  2020, pp 1230--1236, \doi{10.24963/ijcai.2020/171}

\bibitem[{Cheng et~al.(2019)Cheng, Reddy, Saraswat, and
  Lapata}]{cheng-etal-2019-learning}
Cheng J, Reddy S, Saraswat V, Lapata M (2019) Learning an executable neural
  semantic parser. Computational Linguistics pp 59--94,
  \doi{10.1162/coli_a_00342}

\bibitem[{Choi et~al.(2018)Choi, He, Iyyer, Yatskar, Yih, Choi, Liang, and
  Zettlemoyer}]{DBLP:conf/emnlp/ChoiHIYYCLZ18}
Choi E, He H, Iyyer M, Yatskar M, Yih W, Choi Y, Liang P, Zettlemoyer L (2018)
  Qu{AC}: question answering in context. In: Proceedings of the Conference on
  Empirical Methods in Natural Language Processing, {EMNLP} 2018, pp
  2174--2184, \doi{10.18653/v1/d18-1241}

\bibitem[{Christmann et~al.(2019)Christmann, Roy, Abujabal, Singh, and
  Weikum}]{DBLP:conf/cikm/ChristmannRASW19}
Christmann P, Roy RS, Abujabal A, Singh J, Weikum G (2019) Look before you hop:
  conversational question answering over knowledge graphs using judicious
  context expansion. In: Proceedings of the 28th {ACM} International Conference
  on Information and Knowledge Management, {CIKM} 2019, pp 729--738,
  \doi{10.1145/3357384.3358016}

\bibitem[{Chung et~al.(2014)Chung, Gulcehre, Cho, and
  Bengio}]{69e088c8129341ac89810907fe6b1bfe}
Chung J, Gulcehre C, Cho K, Bengio Y (2014) Empirical evaluation of gated
  recurrent neural networks on sequence modeling. In: Proceedings of the 37th
  International Conference on Neural Information Processing Systems, {NIPS}
  2014, pp 01--09

\bibitem[{Cui et~al.(2017{\natexlab{a}})Cui, Huang, Wei, Tan, Duan, and
  Zhou}]{DBLP:conf/acl/CuiHWTDZ17}
Cui L, Huang S, Wei F, Tan C, Duan C, Zhou M (2017{\natexlab{a}}) Superagent: a
  customer service chatbot for e-commerce websites. In: Proceedings of the 55th
  Annual Meeting of the Association for Computational Linguistics, {ACL} 2017,
  pp 97--102, \doi{10.18653/v1/P17-4017}

\bibitem[{Cui et~al.(2017{\natexlab{b}})Cui, Xiao, Wang, Song, Hwang, and
  Wang}]{DBLP:journals/pvldb/CuiXWSHW17}
Cui W, Xiao Y, Wang H, Song Y, Hwang S, Wang W (2017{\natexlab{b}}) {KBQA:}
  learning question answering over {QA} corpora and knowledge bases.
  Proceedings of the {VLDB} Endowment 10(5):565--576,
  \doi{10.14778/3055540.3055549}

\bibitem[{Devlin et~al.(2019)Devlin, Chang, Lee, and
  Toutanova}]{DBLP:conf/naacl/DevlinCLT19}
Devlin J, Chang M, Lee K, Toutanova K (2019) {BERT:} pre-training of deep
  bidirectional transformers for language understanding. In: Proceedings of the
  Conference of the North American Chapter of the Association for Computational
  Linguistics: Human Language Technologies, {NAACL-HLT} 2019, pp 4171--4186,
  \doi{10.18653/v1/n19-1423}

\bibitem[{Dhingra et~al.(2017)Dhingra, Li, Li, Gao, Chen, Ahmed, and
  Deng}]{DBLP:conf/acl/DhingraLLGCAD17}
Dhingra B, Li L, Li X, Gao J, Chen Y, Ahmed F, Deng L (2017) Towards end-to-end
  reinforcement learning of dialogue agents for information access. In:
  Proceedings of the 55th Annual Meeting of the Association for Computational
  Linguistics, {ACL} 2017, pp 484--495, \doi{10.18653/v1/P17-1045}

\bibitem[{Fan et~al.(2018)Fan, Lewis, and Dauphin}]{DBLP:conf/acl/LewisDF18}
Fan A, Lewis M, Dauphin YN (2018) Hierarchical neural story generation. In:
  Proceedings of the 56th Annual Meeting of the Association for Computational
  Linguistics, {ACL} 2018, pp 889--898, \doi{10.18653/v1/P18-1082}

\bibitem[{Fu et~al.(2020)Fu, Qiu, Tang, Li, Yu, and Sun}]{fu2020survey}
Fu B, Qiu Y, Tang C, Li Y, Yu H, Sun J (2020) A survey on complex question
  answering over knowledge base: recent advances and challenges.
  arXiv:200713069

\bibitem[{Gao et~al.(2019)Gao, Galley, and Li}]{DBLP:journals/ftir/GaoGL19}
Gao J, Galley M, Li L (2019) Neural approaches to conversational {AI}.
  Foundations and Trends in Information Retrieval 13(2-3):127--298,
  \doi{10.1561/1500000074}

\bibitem[{Guo et~al.(2018)Guo, Tang, Duan, Zhou, and
  Yin}]{DBLP:conf/nips/GuoTDZY18}
Guo D, Tang D, Duan N, Zhou M, Yin J (2018) Dialog-to-action: conversational
  question answering over a large-scale knowledge base. In: Proceedings of the
  32nd International Conference on Neural Information Processing Systems, NIPS
  2018, pp 2946--2955

\bibitem[{Gupta et~al.(2020)Gupta, Rawat, and
  Yu}]{gupta-etal-2020-conversational}
Gupta S, Rawat BPS, Yu H (2020) Conversational machine comprehension: a
  literature review. In: Proceedings of the 28th International Conference on
  Computational Linguistics, {COLING} 2020, pp 2739--2753,
  \doi{10.18653/v1/2020.coling-main.247}

\bibitem[{Gur et~al.(2017)Gur, Hewlett, Lacoste, and
  Jones}]{DBLP:conf/emnlp/HewlettJLG17}
Gur I, Hewlett D, Lacoste A, Jones L (2017) Accurate supervised and
  semi-supervised machine reading for long documents. In: Proceedings of the
  Conference on Empirical Methods in Natural Language Processing, {EMNLP} 2017,
  pp 2011--2020, \doi{10.18653/v1/d17-1214}

\bibitem[{Hermann et~al.(2015)Hermann, Kocisk{\'{y}}, Grefenstette, Espeholt,
  Kay, Suleyman, and Blunsom}]{DBLP:conf/nips/HermannKGEKSB15}
Hermann KM, Kocisk{\'{y}} T, Grefenstette E, Espeholt L, Kay W, Suleyman M,
  Blunsom P (2015) Teaching machines to read and comprehend. In: Proceedings of
  the 28th International Conference on Neural Information Processing Systems,
  {NIPS} 2015, pp 1693--1701

\bibitem[{Higashinaka and Isozaki(2008)}]{higashinaka-isozaki-2008-corpus}
Higashinaka R, Isozaki H (2008) Corpus-based question answering for
  why-questions. In: Proceedings of the 3rd International Joint Conference on
  Natural Language Processing, {IJCNLP} 2008, pp 01--08

\bibitem[{Hochreiter and Schmidhuber(1997)}]{hochreiter1997long}
Hochreiter S, Schmidhuber J (1997) Long short-term memory. Neural Computation
  9(8):1735--1780

\bibitem[{Huang et~al.(2019{\natexlab{a}})Huang, Choi, and
  Yih}]{DBLP:conf/iclr/HuangCY19}
Huang H, Choi E, Yih W (2019{\natexlab{a}}) Flow{QA}: grasping flow in history
  for conversational machine comprehension. In: Proceedings of the 7th
  International Conference on Learning Representations, {ICLR} 2019, pp 01--08

\bibitem[{Huang et~al.(2019{\natexlab{b}})Huang, Bras, Bhagavatula, and
  Choi}]{DBLP:conf/emnlp/HuangBBC19}
Huang L, Bras RL, Bhagavatula C, Choi Y (2019{\natexlab{b}}) Cosmos {QA:}
  machine reading comprehension with contextual commonsense reasoning. In:
  Proceedings of the Conference on Empirical Methods in Natural Language
  Processing and the 9th International Joint Conference on Natural Language
  Processing, {EMNLP-IJCNLP} 2019, pp 2391--2401, \doi{10.18653/v1/D19-1243}

\bibitem[{Iyyer et~al.(2014)Iyyer, Boyd{-}Graber, Claudino, Socher, and
  III}]{DBLP:conf/emnlp/IyyerBCSD14}
Iyyer M, Boyd{-}Graber JL, Claudino LMB, Socher R, III HD (2014) A neural
  network for factoid question answering over paragraphs. In: Proceedings of
  the Conference on Empirical Methods in Natural Language Processing, {EMNLP}
  2014, pp 633--644, \doi{10.3115/v1/d14-1070}

\bibitem[{Iyyer et~al.(2017)Iyyer, Yih, and Chang}]{DBLP:conf/acl/IyyerYC17}
Iyyer M, Yih W, Chang M (2017) Search-based neural structured learning for
  sequential question answering. In: Proceedings of the 55th Annual Meeting of
  the Association for Computational Linguistics, {ACL} 2017, pp 1821--1831,
  \doi{10.18653/v1/P17-1167}

\bibitem[{Jiang et~al.(2019)Jiang, Wu, and Jiang}]{jiang-etal-2019-freebaseqa}
Jiang K, Wu D, Jiang H (2019) {F}reebase{QA}: A new factoid {QA} data set
  matching trivia-style question-answer pairs with {F}reebase. In: Proceedings
  of the Conference of the North {A}merican Chapter of the Association for
  Computational Linguistics: Human Language Technologies, {NAACL-HLT} 2019, pp
  318--323, \doi{10.18653/v1/N19-1028}

\bibitem[{Joshi et~al.(2017)Joshi, Choi, Weld, and
  Zettlemoyer}]{DBLP:conf/acl/JoshiCWZ17}
Joshi M, Choi E, Weld DS, Zettlemoyer L (2017) Trivia{QA}: a large scale
  distantly supervised challenge dataset for reading comprehension. In:
  Proceedings of the 55th Annual Meeting of the Association for Computational
  Linguistics, {ACL} 2017, pp 1601--1611, \doi{10.18653/v1/P17-1147}

\bibitem[{Kacupaj et~al.(2021)Kacupaj, Plepi, Singh, Thakkar, Lehmann, and
  Maleshkova}]{kacupaj-etal-2021-conversational}
Kacupaj E, Plepi J, Singh K, Thakkar H, Lehmann J, Maleshkova M (2021)
  Conversational question answering over knowledge graphs with transformer and
  graph attention networks. In: Proceedings of the 16th Conference of the
  European Chapter of the Association for Computational Linguistics, {EACL}
  2021, pp 850--862

\bibitem[{Kocisk{\'{y}} et~al.(2018)Kocisk{\'{y}}, Schwarz, Blunsom, Dyer,
  Hermann, Melis, and Grefenstette}]{DBLP:journals/tacl/KociskySBDHMG18}
Kocisk{\'{y}} T, Schwarz J, Blunsom P, Dyer C, Hermann KM, Melis G,
  Grefenstette E (2018) The narrative{QA} reading comprehension challenge.
  Transactions of the Association for Computational Linguistics, {ACL} 2018, pp
  317--328

\bibitem[{Lai et~al.(2017)Lai, Xie, Liu, Yang, and
  Hovy}]{DBLP:conf/emnlp/LaiXLYH17}
Lai G, Xie Q, Liu H, Yang Y, Hovy EH (2017) {RACE:} large-scale reading
  comprehension dataset from examinations. In: Proceedings of the Conference on
  Empirical Methods in Natural Language Processing, {EMNLP} 2017, pp 785--794,
  \doi{10.18653/v1/d17-1082}

\bibitem[{Lan et~al.(2020)Lan, Chen, Goodman, Gimpel, Sharma, and
  Soricut}]{DBLP:conf/iclr/LanCGGSS20}
Lan Z, Chen M, Goodman S, Gimpel K, Sharma P, Soricut R (2020) {ALBERT:} {A}
  lite {BERT} for self-supervised learning of language representations. In:
  Proceedings of the 8th International Conference on Learning Representations,
  {ICLR} 2020

\bibitem[{Lehmann et~al.(2015)Lehmann, Isele, Jakob, Jentzsch, Kontokostas,
  Mendes, Hellmann, Morsey, Van~Kleef, Auer et~al.}]{lehmann2015dbpedia}
Lehmann J, Isele R, Jakob M, Jentzsch A, Kontokostas D, Mendes PN, Hellmann S,
  Morsey M, Van~Kleef P, Auer S, et~al. (2015) {DB}pedia--a large-scale,
  multilingual knowledge base extracted from wikipedia. Semantic Web
  6(2):167--195

\bibitem[{Lin et~al.(2019)Lin, Chen, Chen, and Ren}]{DBLP:conf/emnlp/LinCCR19}
Lin BY, Chen X, Chen J, Ren X (2019) Kag{N}et: knowledge-aware graph networks
  for commonsense reasoning. In: Proceedings of the Conference on Empirical
  Methods in Natural Language Processing and the 9th International Joint
  Conference on Natural Language Processing, {EMNLP-IJCNLP} 2019, pp
  2829--2839, \doi{10.18653/v1/D19-1282}

\bibitem[{{Liu} et~al.(2019){Liu}, {Zhang}, {Zhang}, and {Wang}}]{8651505}
{Liu} S, {Zhang} S, {Zhang} X, {Wang} H (2019) R-{T}rans: {RNN} transformer
  network for chinese machine reading comprehension. IEEE Access
  7:27736--27745, \doi{10.1109/ACCESS.2019.2901547}

\bibitem[{Liu et~al.(2019{\natexlab{a}})Liu, Zhang, Zhang, Wang, and
  Zhang}]{liu2019neural}
Liu S, Zhang X, Zhang S, Wang H, Zhang W (2019{\natexlab{a}}) Neural machine
  reading comprehension: methods and trends. Applied Sciences 9(18):3698

\bibitem[{Liu et~al.(2019{\natexlab{b}})Liu, Ott, Goyal, Du, Joshi, Chen, Levy,
  Lewis, Zettlemoyer, and Stoyanov}]{liu2019roberta}
Liu Y, Ott M, Goyal N, Du J, Joshi M, Chen D, Levy O, Lewis M, Zettlemoyer L,
  Stoyanov V (2019{\natexlab{b}}) Ro{BERT}a: a robustly optimized bert
  pretraining approach. arXiv:190711692

\bibitem[{Lu et~al.(2019)Lu, Pramanik, Roy, Abujabal, Wang, and
  Weikum}]{DBLP:conf/sigir/LuPRAWW19}
Lu X, Pramanik S, Roy RS, Abujabal A, Wang Y, Weikum G (2019) Answering complex
  questions by joining multi-document evidence with quasi knowledge graphs. In:
  Proceedings of the 42nd International Conference on Research and Development
  in Information Retrieval, {SIGIR} 2019, pp 105--114,
  \doi{10.1145/3331184.3331252}

\bibitem[{Luong et~al.(2015)Luong, Pham, and
  Manning}]{luong-etal-2015-effective}
Luong T, Pham H, Manning CD (2015) Effective approaches to attention-based
  neural machine translation. In: Proceedings of the Conference on Empirical
  Methods in Natural Language Processing, {EMNLP} 2015, pp 1412--1421,
  \doi{10.18653/v1/D15-1166}

\bibitem[{Lv et~al.(2020)Lv, Guo, Xu, Tang, Duan, Gong, Shou, Jiang, Cao, and
  Hu}]{DBLP:conf/aaai/LvGXTDGSJCH20}
Lv S, Guo D, Xu J, Tang D, Duan N, Gong M, Shou L, Jiang D, Cao G, Hu S (2020)
  Graph-based reasoning over heterogeneous external knowledge for commonsense
  question answering. In: Proceeding of the 34th Conference on Artificial
  Intelligence, {AAAI} 2020, pp 8449--8456

\bibitem[{Martinez-Gil(2015)}]{martinez2015automated}
Martinez-Gil J (2015) Automated knowledge base management: A survey. Computer
  Science Review 18:1--9

\bibitem[{McCann et~al.(2017)McCann, Bradbury, Xiong, and
  Socher}]{NIPS2017_7209}
McCann B, Bradbury J, Xiong C, Socher R (2017) Learned in translation:
  contextualized word vectors. In: Proceedings of the 31st International
  Conference on Neural Information Processing Systems, {NIPS} 2017, pp
  6294--6305

\bibitem[{Mikolov et~al.(2013)Mikolov, Chen, Corrado, and
  Dean}]{DBLP:journals/corr/abs-1301-3781}
Mikolov T, Chen K, Corrado G, Dean J (2013) Efficient estimation of word
  representations in vector space. In: Proceedings of the 1st International
  Conference on Learning Representations, {ICLR} 2013, pp 01--12

\bibitem[{Miller et~al.(2016)Miller, Fisch, Dodge, Karimi, Bordes, and
  Weston}]{miller-etal-2016-key}
Miller A, Fisch A, Dodge J, Karimi AH, Bordes A, Weston J (2016) Key-value
  memory networks for directly reading documents. In: Proceedings of the
  Conference on Empirical Methods in Natural Language Processing, {EMNLP} 2016,
  pp 1400--1409, \doi{10.18653/v1/D16-1147}

\bibitem[{Miller et~al.(2017)Miller, Feng, Batra, Bordes, Fisch, Lu, Parikh,
  and Weston}]{DBLP:conf/emnlp/MillerFBBFLPW17}
Miller AH, Feng W, Batra D, Bordes A, Fisch A, Lu J, Parikh D, Weston J (2017)
  Parl{AI}: a dialog research software platform. In: Proceedings of the
  Conference on Empirical Methods in Natural Language Processing, {EMNLP} 2017,
  pp 79--84, \doi{10.18653/v1/d17-2014}

\bibitem[{Mishra and Jain(2016)}]{mishra2016survey}
Mishra A, Jain SK (2016) A survey on question answering systems with
  classification. Journal of King Saud University-Computer and Information
  Sciences 28(3):345--361

\bibitem[{Mitchell et~al.(2018)Mitchell, Cohen, Hruschka, Talukdar, Yang,
  Betteridge, Carlson, Dalvi, Gardner, Kisiel et~al.}]{mitchell2018never}
Mitchell T, Cohen W, Hruschka E, Talukdar P, Yang B, Betteridge J, Carlson A,
  Dalvi B, Gardner M, Kisiel B, et~al. (2018) Never-ending learning.
  Communications of the ACM 61(5):103--115

\bibitem[{Monz(2011)}]{monz2011machine}
Monz C (2011) Machine learning for query formulation in question answering.
  Natural Language Engineering 17(4):425

\bibitem[{M{\"{u}}ller et~al.(2019)M{\"{u}}ller, Piccinno, Shaw, Nicosia, and
  Altun}]{DBLP:conf/emnlp/MullerPSNA19}
M{\"{u}}ller T, Piccinno F, Shaw P, Nicosia M, Altun Y (2019) Answering
  conversational questions on structured data without logical forms. In:
  Proceedings of the Conference on Empirical Methods in Natural Language
  Processing and the 9th International Joint Conference on Natural Language
  Processing, {EMNLP-IJCNLP} 2019, pp 5901--5909, \doi{10.18653/v1/D19-1603}

\bibitem[{Nallapati et~al.(2016)Nallapati, Zhou, dos Santos,
  G{\"{u}}l{\c{c}}ehre, and Xiang}]{DBLP:conf/conll/NallapatiZSGX16}
Nallapati R, Zhou B, dos Santos CN, G{\"{u}}l{\c{c}}ehre {\c{C}}, Xiang B
  (2016) Abstractive text summarization using sequence-to-sequence rnns and
  beyond. In: Proceedings of the 20th {SIGNLL} Conference on Computational
  Natural Language Learning, CoNLL 2016, pp 280--290,
  \doi{10.18653/v1/k16-1028}

\bibitem[{Nguyen et~al.(2016)Nguyen, Rosenberg, Song, Gao, Tiwary, Majumder,
  and Deng}]{DBLP:conf/nips/NguyenRSGTMD16}
Nguyen T, Rosenberg M, Song X, Gao J, Tiwary S, Majumder R, Deng L (2016) {MS}
  {MARCO:} a human generated machine reading comprehension dataset. In:
  Proceedings of the 30th Annual Conference on Neural Information Processing
  Systems, {NIPS} 2016, pp 01--11

\bibitem[{Ohsugi et~al.(2019)Ohsugi, Saito, Nishida, Asano, and
  Tomita}]{ohsugi-etal-2019-simple}
Ohsugi Y, Saito I, Nishida K, Asano H, Tomita J (2019) A simple but effective
  method to incorporate multi-turn context with {BERT} for conversational
  machine comprehension. In: Proceedings of the 57th Annual Meeting of the
  Association for Computational Linguistics, {ACL} 2019, pp 11--17,
  \doi{10.18653/v1/W19-4102}

\bibitem[{Ostermann et~al.(2018)Ostermann, Modi, Roth, Thater, and
  Pinkal}]{DBLP:conf/lrec/0002MRTP18}
Ostermann S, Modi A, Roth M, Thater S, Pinkal M (2018) M{CS}cript: a novel
  dataset for assessing machine comprehension using script knowledge. In:
  Proceedings of the 11th International Conference on Language Resources and
  Evaluation, {LREC} 2018, pp 01--08

\bibitem[{Pasupat and Liang(2015)}]{pasupat-liang-2015-compositional}
Pasupat P, Liang P (2015) Compositional semantic parsing on semi-structured
  tables. In: Proceedings of the 53rd Annual Meeting of the Association for
  Computational Linguistics and the 7th International Joint Conference on
  Natural Language Processing, {IJCNLP} 2015, pp 1470--1480,
  \doi{10.3115/v1/P15-1142}

\bibitem[{Peng et~al.(2020)Peng, Zhu, Li, Li, Li, Zeng, and
  Gao}]{DBLP:conf/emnlp/PengZLLLZG20}
Peng B, Zhu C, Li C, Li X, Li J, Zeng M, Gao J (2020) Few-shot natural language
  generation for task-oriented dialog. In: Proceedings of the Conference on
  Empirical Methods in Natural Language Processing, {EMNLP} 2020, pp 172--182,
  \doi{10.18653/v1/2020.findings-emnlp.17}

\bibitem[{Pennington et~al.(2014)Pennington, Socher, and
  Manning}]{pennington-etal-2014-glove}
Pennington J, Socher R, Manning C (2014) {G}lo{V}e: global vectors for word
  representation. In: Proceedings of the Conference on Empirical Methods in
  Natural Language Processing, {EMNLP} 2014, pp 1532--1543,
  \doi{10.3115/v1/D14-1162}

\bibitem[{Peters et~al.(2018)Peters, Neumann, Iyyer, Gardner, Clark, Lee, and
  Zettlemoyer}]{peters-etal-2018-deep}
Peters M, Neumann M, Iyyer M, Gardner M, Clark C, Lee K, Zettlemoyer L (2018)
  Deep contextualized word representations. In: Proceedings of the Conference
  of the North {A}merican Chapter of the Association for Computational
  Linguistics: Human Language Technologies, {NAACL-HLT} 2018, pp 2227--2237,
  \doi{10.18653/v1/N18-1202}

\bibitem[{Pinto et~al.(2002)Pinto, Branstein, Coleman, Croft, King, Li, and
  Wei}]{pinto2002quasm}
Pinto D, Branstein M, Coleman R, Croft WB, King M, Li W, Wei X (2002) Quasm: A
  system for question answering using semi-structured data. In: Proceedings of
  the 2nd ACM/IEEE-CS joint conference on Digital libraries, pp 46--55

\bibitem[{Qi et~al.(2019)Qi, Lin, Mehr, Wang, and
  Manning}]{DBLP:conf/emnlp/QiLMWM19}
Qi P, Lin X, Mehr L, Wang Z, Manning CD (2019) Answering complex open-domain
  questions through iterative query generation. In: Proceedings of the
  Conference on Empirical Methods in Natural Language Processing and the 9th
  International Joint Conference on Natural Language Processing, {EMNLP-IJCNLP}
  2019, pp 2590--2602, \doi{10.18653/v1/D19-1261}

\bibitem[{Qiu et~al.(2021)Qiu, Huang, Chen, Ji, Qu, Wei, Huang, and
  Zhang}]{qiu2021reinforced}
Qiu M, Huang X, Chen C, Ji F, Qu C, Wei W, Huang J, Zhang Y (2021) Reinforced
  history backtracking for conversational question answering. In: Proceedings
  of the 35th Conference on Artificial Intelligence, {AAAI} 2021

\bibitem[{Qu et~al.(2019{\natexlab{a}})Qu, Yang, Qiu, Croft, Zhang, and
  Iyyer}]{DBLP:conf/sigir/Qu0QCZI19}
Qu C, Yang L, Qiu M, Croft WB, Zhang Y, Iyyer M (2019{\natexlab{a}}) {BERT}
  with history answer embedding for conversational question answering. In:
  Proceedings of the 42nd International Conference on Research and Development
  in Information Retrieval, {SIGIR} 2019, pp 1133--1136,
  \doi{10.1145/3331184.3331341}

\bibitem[{Qu et~al.(2019{\natexlab{b}})Qu, Yang, Qiu, Zhang, Chen, Croft, and
  Iyyer}]{DBLP:conf/cikm/QuYQZCCI19}
Qu C, Yang L, Qiu M, Zhang Y, Chen C, Croft WB, Iyyer M (2019{\natexlab{b}})
  Attentive history selection for conversational question answering. In:
  Proceedings of the 28th International Conference on Information and Knowledge
  Management, {CIKM} 2019, pp 1391--1400, \doi{10.1145/3357384.3357905}

\bibitem[{Qu et~al.(2020)Qu, Yang, Chen, Qiu, Croft, and Iyyer}]{qu2020open}
Qu C, Yang L, Chen C, Qiu M, Croft WB, Iyyer M (2020) Open-retrieval
  conversational question answering. In: Proceedings of the 43rd International
  Conference on Research and Development in Information Retrieval, pp 539--548

\bibitem[{Qu et~al.(2021)Qu, Yang, Chen, Croft, Krishna, and
  Iyyer}]{DBLP:conf/ecir/QuYCCKI21}
Qu C, Yang L, Chen C, Croft WB, Krishna K, Iyyer M (2021) Weakly-supervised
  open-retrieval conversational question answering. In: Proceedings of the 43rd
  European Conference on {IR} Research, {ECIR} 2021, pp 529--543,
  \doi{10.1007/978-3-030-72113-8\_35}

\bibitem[{Radford et~al.(2018)Radford, Narasimhan, Salimans, and
  Sutskever}]{radford2018improving}
Radford A, Narasimhan K, Salimans T, Sutskever I (2018) Improving language
  understanding by generative pre-training

\bibitem[{Radford et~al.(2019)Radford, Wu, Child, Luan, Amodei, and
  Sutskever}]{radford2019language}
Radford A, Wu J, Child R, Luan D, Amodei D, Sutskever I (2019) Language models
  are unsupervised multitask learners. OpenAI Blog 1(8):9

\bibitem[{Rajani et~al.(2019)Rajani, McCann, Xiong, and
  Socher}]{DBLP:conf/acl/RajaniMXS19}
Rajani NF, McCann B, Xiong C, Socher R (2019) Explain yourself! leveraging
  language models for commonsense reasoning. In: Proceedings of the 57th
  Conference of the Association for Computational Linguistics, {ACL} 2019, pp
  4932--4942, \doi{10.18653/v1/p19-1487}

\bibitem[{Rajpurkar et~al.(2016)Rajpurkar, Zhang, Lopyrev, and
  Liang}]{DBLP:conf/emnlp/RajpurkarZLL16}
Rajpurkar P, Zhang J, Lopyrev K, Liang P (2016) S{Q}u{AD}: 100, 000+ questions
  for machine comprehension of text. In: Proceedings of the Conference on
  Empirical Methods in Natural Language Processing, {EMNLP} 2016, pp
  2383--2392, \doi{10.18653/v1/d16-1264}

\bibitem[{Rajpurkar et~al.(2018)Rajpurkar, Jia, and
  Liang}]{rajpurkar-etal-2018-know}
Rajpurkar P, Jia R, Liang P (2018) Know what you don{'}t know: Unanswerable
  questions for {SQ}u{AD}. In: Proceedings of the 56th Annual Meeting of the
  Association for Computational Linguistics, {ACL} 2018, pp 784--789,
  \doi{10.18653/v1/P18-2124}

\bibitem[{Reddy et~al.(2019)Reddy, Chen, and
  Manning}]{DBLP:journals/tacl/ReddyCM19}
Reddy S, Chen D, Manning CD (2019) Co{QA}: a conversational question answering
  challenge. Transactions of the Association for Computational Linguistics
  7:249--266

\bibitem[{Ren et~al.(2018)Ren, Xie, Chen, and Yu}]{DBLP:conf/emnlp/RenXCY18}
Ren L, Xie K, Chen L, Yu K (2018) Towards universal dialogue state tracking.
  In: Proceedings of the Conference on Empirical Methods in Natural Language
  Processing, {EMNLP} 2019, pp 2780--2786, \doi{10.18653/v1/d18-1299}

\bibitem[{Richardson et~al.(2013)Richardson, Burges, and
  Renshaw}]{DBLP:conf/emnlp/RichardsonBR13}
Richardson M, Burges CJC, Renshaw E (2013) M{C}{T}est: a challenge dataset for
  the open-domain machine comprehension of text. In: Proceedings of the
  Conference on Empirical Methods in Natural Language Processing, {EMNLP} 2013,
  pp 193--203

\bibitem[{Rumelhart et~al.(1986)Rumelhart, Hinton, and
  Williams}]{rumelhart1986learning}
Rumelhart DE, Hinton GE, Williams RJ (1986) Learning representations by
  back-propagating errors. nature 323(6088):533--536

\bibitem[{Saeidi et~al.(2018)Saeidi, Bartolo, Lewis, Singh, Rockt{\"{a}}schel,
  Sheldon, Bouchard, and Riedel}]{DBLP:conf/emnlp/SaeidiBL0RSB018}
Saeidi M, Bartolo M, Lewis PSH, Singh S, Rockt{\"{a}}schel T, Sheldon M,
  Bouchard G, Riedel S (2018) Interpretation of natural language rules in
  conversational machine reading. In: Proceedings of the Conference on
  Empirical Methods in Natural Language Processing, {EMNLP} 2018, pp
  2087--2097, \doi{10.18653/v1/d18-1233}

\bibitem[{Saha et~al.(2018)Saha, Pahuja, Khapra, Sankaranarayanan, and
  Chandar}]{DBLP:conf/aaai/SahaPKSC18}
Saha A, Pahuja V, Khapra MM, Sankaranarayanan K, Chandar S (2018) Complex
  sequential question answering: towards learning to converse over linked
  question answer pairs with a knowledge graph. In: Proceedings of the 32nd
  Conference on Artificial Intelligence, {AAAI} 2018, pp 705--713

\bibitem[{Sap et~al.(2019)Sap, Le~Bras, Allaway, Bhagavatula, Lourie, Rashkin,
  Roof, Smith, and Choi}]{sap2019atomic}
Sap M, Le~Bras R, Allaway E, Bhagavatula C, Lourie N, Rashkin H, Roof B, Smith
  NA, Choi Y (2019) {ATOMIC}: an atlas of machine commonsense for if-then
  reasoning. In: Proceedings of the 33rd Conference on Artificial Intelligence,
  {AAAI} 2019, vol~33, pp 3027--3035

\bibitem[{Scarselli et~al.(2008)Scarselli, Gori, Tsoi, Hagenbuchner, and
  Monfardini}]{scarselli2008graph}
Scarselli F, Gori M, Tsoi AC, Hagenbuchner M, Monfardini G (2008) The graph
  neural network model. IEEE Transactions on Neural Networks 20(1):61--80

\bibitem[{See et~al.(2017)See, Liu, and Manning}]{see-etal-2017-get}
See A, Liu PJ, Manning CD (2017) Get to the point: summarization with
  pointer-generator networks. In: Proceedings of the 55th Annual Meeting of the
  Association for Computational Linguistics, {ACL} 2017, pp 1073--1083,
  \doi{10.18653/v1/P17-1099}

\bibitem[{Seo et~al.(2017)Seo, Kembhavi, Farhadi, and
  Hajishirzi}]{DBLP:conf/iclr/SeoKFH17}
Seo MJ, Kembhavi A, Farhadi A, Hajishirzi H (2017) Bidirectional attention flow
  for machine comprehension. In: 5th International Conference on Learning
  Representations, {ICLR} 2017, pp 01--13

\bibitem[{Serban et~al.(2016)Serban, Sordoni, Bengio, Courville, and
  Pineau}]{serban2016building}
Serban I, Sordoni A, Bengio Y, Courville A, Pineau J (2016) Building end-to-end
  dialogue systems using generative hierarchical neural network models. In:
  Proceedings of the 30th Conference on Artificial Intelligence, {AAAI} 2016,
  pp 3776--3784

\bibitem[{Sharma and Goolsbey(2019)}]{DBLP:conf/aaai/SharmaG19}
Sharma A, Goolsbey KM (2019) Simulation-based approach to efficient commonsense
  reasoning in very large knowledge bases. In: Proceedings of the 33rd
  Conference on Artificial Intelligence, {AAAI} 2019, pp 1360--1367,
  \doi{10.1609/aaai.v33i01.33011360}

\bibitem[{Shen and Klakow(2006)}]{DBLP:conf/acl/ShenK06}
Shen D, Klakow D (2006) Exploring correlation of dependency relation paths for
  answer extraction. In: Proceedings of the 44th Annual Meeting of the
  Association for Computational Linguistics, {ACL} 2006, p 889–896,
  \doi{10.3115/1220175.1220287}

\bibitem[{Shen et~al.(2019)Shen, Geng, Qin, Guo, Tang, Duan, Long, and
  Jiang}]{DBLP:conf/emnlp/ShenGQGTDLJ19}
Shen T, Geng X, Qin T, Guo D, Tang D, Duan N, Long G, Jiang D (2019) Multi-task
  learning for conversational question answering over a large-scale knowledge
  base. In: Proceedings of the Conference on Empirical Methods in Natural
  Language Processing and the 9th International Joint Conference on Natural
  Language Processing, {EMNLP-IJCNLP} 2019, pp 2442--2451,
  \doi{10.18653/v1/D19-1248}

\bibitem[{Speer et~al.(2017)Speer, Chin, and Havasi}]{speer2017conceptnet}
Speer R, Chin J, Havasi C (2017) Concept{N}et 5.5: An open multilingual graph
  of general knowledge. In: Proceedings of the 31st Conference on Artificial
  Intelligence, {AAAI} 2017, pp 4444--4451

\bibitem[{Suhr et~al.(2018)Suhr, Iyer, and Artzi}]{suhr-etal-2018-learning}
Suhr A, Iyer S, Artzi Y (2018) Learning to map context-dependent sentences to
  executable formal queries. In: Proceedings of the Conference of the North
  {A}merican Chapter of the Association for Computational Linguistics: Human
  Language Technologies, {NAACL-HLT} 2018, pp 2238--2249,
  \doi{10.18653/v1/N18-1203}

\bibitem[{Sun et~al.(2020)Sun, Cao, Zhao, Wan, Zhou, Zhang, Wang, and
  Zheng}]{sun2020multi}
Sun R, Cao X, Zhao Y, Wan J, Zhou K, Zhang F, Wang Z, Zheng K (2020)
  Multi-modal knowledge graphs for recommender systems. In: Proceedings of the
  29th ACM International Conference on Information \& Knowledge Management,
  {CIKM} 2020, pp 1405--1414

\bibitem[{Suster and Daelemans(2018)}]{DBLP:conf/naacl/SusterD18}
Suster S, Daelemans W (2018) Cli{CR}: a dataset of clinical case reports for
  machine reading comprehension. In: Proceedings of the Conference of the North
  American Chapter of the Association for Computational Linguistics: Human
  Language Technologies, {NAACL-HLT} 2018, pp 1551--1563,
  \doi{10.18653/v1/n18-1140}

\bibitem[{Sutskever et~al.(2014)Sutskever, Vinyals, and Le}]{NIPS2014_a14ac55a}
Sutskever I, Vinyals O, Le QV (2014) Sequence to sequence learning with neural
  networks. In: Advances in Neural Information Processing Systems, vol~27, p
  3104–3112

\bibitem[{Tian et~al.(2017)Tian, Yan, Mou, Song, Feng, and
  Zhao}]{tian-etal-2017-make}
Tian Z, Yan R, Mou L, Song Y, Feng Y, Zhao D (2017) How to make context more
  useful? an empirical study on context-aware neural conversational models. In:
  Proceedings of the 55th Annual Meeting of the Association for Computational
  Linguistics, {ACL} 2017, pp 231--236, \doi{10.18653/v1/P17-2036}

\bibitem[{Trischler et~al.(2017)Trischler, Wang, Yuan, Harris, Sordoni,
  Bachman, and Suleman}]{DBLP:conf/rep4nlp/TrischlerWYHSBS17}
Trischler A, Wang T, Yuan X, Harris J, Sordoni A, Bachman P, Suleman K (2017)
  News{QA}: a machine comprehension dataset. In: Proceedings of the 55th Annual
  Meeting of the Association for Computational Linguistics, {ACL} 2017, pp
  191--200, \doi{10.18653/v1/w17-2623}

\bibitem[{Trivedi et~al.(2017)Trivedi, Maheshwari, Dubey, and
  Lehmann}]{DBLP:conf/semweb/TrivediMDL17}
Trivedi P, Maheshwari G, Dubey M, Lehmann J (2017) L{C}-{Qu}a{D}: a corpus for
  complex question answering over knowledge graphs. In: Proceedings of the 16th
  International Semantic Web Conference, {ISWC} 2017, pp 210--218

\bibitem[{Vaswani et~al.(2017)Vaswani, Shazeer, Parmar, Uszkoreit, Jones,
  Gomez, Kaiser, and Polosukhin}]{NIPS2017_3f5ee243}
Vaswani A, Shazeer N, Parmar N, Uszkoreit J, Jones L, Gomez AN, Kaiser Lu,
  Polosukhin I (2017) Attention is all you need. In: Proceedings of the 31st
  International Conference on Neural Information Processing Systems, {NIPS}
  2017, vol~30, pp 5998–6008,

\bibitem[{Velickovic et~al.(2018)Velickovic, Cucurull, Casanova, Romero,
  Li{\`{o}}, and Bengio}]{DBLP:conf/iclr/VelickovicCCRLB18}
Velickovic P, Cucurull G, Casanova A, Romero A, Li{\`{o}} P, Bengio Y (2018)
  Graph attention networks. In: Proceedings of the 6th International Conference
  on Learning Representations, {ICLR} 2018,

\bibitem[{Vinyals et~al.(2015)Vinyals, Fortunato, and
  Jaitly}]{NIPS2015_29921001}
Vinyals O, Fortunato M, Jaitly N (2015) Pointer networks. In: Proceedings of
  the 29th International Conference on Neural Information Processing Systems,
  {NIPS} 2015, vol~28, pp 2692--2700

\bibitem[{Wang et~al.(2018)Wang, Zhang, Ma, Sun, Wang, and
  Wang}]{wang-etal-2018-neural-question}
Wang H, Zhang X, Ma S, Sun X, Wang H, Wang M (2018) A neural question answering
  model based on semi-structured tables. In: Proceedings of the 27th
  International Conference on Computational Linguistics, {COLING} 2018, pp
  1941--1951

\bibitem[{Welbl et~al.(2017)Welbl, Liu, and
  Gardner}]{DBLP:conf/aclnut/WelblLG17}
Welbl J, Liu NF, Gardner M (2017) Crowdsourcing multiple choice science
  questions. In: Proceedings of the Conference on Empirical Methods in Natural
  Language Processing, {EMNLP} 2017, pp 94--106, \doi{10.18653/v1/w17-4413}

\bibitem[{Wen et~al.(2017)Wen, Vandyke, Mrk{\v{s}}i{\'c}, Ga{\v{s}}i{\'c},
  Rojas-Barahona, Su, Ultes, and Young}]{wen-etal-2017-network}
Wen TH, Vandyke D, Mrk{\v{s}}i{\'c} N, Ga{\v{s}}i{\'c} M, Rojas-Barahona LM, Su
  PH, Ultes S, Young S (2017) A network-based end-to-end trainable
  task-oriented dialogue system. In: Proceedings of the 15th Conference of the
  {E}uropean Chapter of the Association for Computational Linguistics, {EACL}
  2017, pp 438--449

\bibitem[{Wu et~al.(2015)Wu, Li, and Lee}]{wu2015probabilistic}
Wu J, Li M, Lee CH (2015) A probabilistic framework for representing dialog
  systems and entropy-based dialog management through dynamic stochastic state
  evolution. IEEE/ACM Transactions on Audio, Speech, and Language Processing
  23(11):2026--2035

\bibitem[{Wu et~al.(2016)Wu, Schuster, Chen, Le, Norouzi, Macherey, Krikun,
  Cao, Gao, Macherey, Klingner, Shah, Johnson, Liu, Kaiser, Gouws, Kato, Kudo,
  Kazawa, Stevens, Kurian, Patil, Wang, Young, Smith, Riesa, Rudnick, Vinyals,
  Corrado, Hughes, and Dean}]{DBLP:journals/corr/WuSCLNMKCGMKSJL16}
Wu Y, Schuster M, Chen Z, Le QV, Norouzi M, Macherey W, Krikun M, Cao Y, Gao Q,
  Macherey K, Klingner J, Shah A, Johnson M, Liu X, Kaiser L, Gouws S, Kato Y,
  Kudo T, Kazawa H, Stevens K, Kurian G, Patil N, Wang W, Young C, Smith J,
  Riesa J, Rudnick A, Vinyals O, Corrado G, Hughes M, Dean J (2016) Google's
  neural machine translation system: Bridging the gap between human and machine
  translation. arXiv:160908144

\bibitem[{Xiong et~al.(2021)Xiong, Li, Iyer, Du, Lewis, Wang, Mehdad, Yih,
  Riedel, Kiela, and Oguz}]{xiong2021answering}
Xiong W, Li X, Iyer S, Du J, Lewis P, Wang WY, Mehdad Y, Yih S, Riedel S, Kiela
  D, Oguz B (2021) Answering complex open-domain questions with multi-hop dense
  retrieval. In: Proceedings of the 9th International Conference on Learning
  Representations, {ICLR} 2021, pp 01 --19

\bibitem[{Yang et~al.(2015)Yang, Yih, and Meek}]{yang-etal-2015-wikiqa}
Yang Y, Yih Wt, Meek C (2015) {W}iki{QA}: A challenge dataset for open-domain
  question answering. In: Proceedings of the Conference on Empirical Methods in
  Natural Language Processing, {EMNLP} 2015, pp 2013--2018,
  \doi{10.18653/v1/D15-1237}

\bibitem[{Yang et~al.(2019)Yang, Dai, Yang, Carbonell, Salakhutdinov, and
  Le}]{NEURIPS2019_dc6a7e65}
Yang Z, Dai Z, Yang Y, Carbonell J, Salakhutdinov RR, Le QV (2019) X{LN}et:
  generalized autoregressive pretraining for language understanding. In:
  Proceedings of the 33rd International Conference on Neural Information
  Processing Systems, {NIPS} 2019, vol~32, pp 5754--5764

\bibitem[{Yatskar(2019)}]{DBLP:conf/naacl/Yatskar19}
Yatskar M (2019) A qualitative comparison of {CoQA}, {SQuAD} 2.0 and {QuAC}.
  In: Proceedings of the Conference of the North American Chapter of the
  Association for Computational Linguistics: Human Language Technologies,
  {NAACL-HLT} 2019, pp 2318--2323, \doi{10.18653/v1/n19-1241}

\bibitem[{Yeh and Chen(2019)}]{DBLP:conf/acl-mrqa/YehC19}
Yeh Y, Chen Y (2019) Flow{D}elta: modeling flow information gain in reasoning
  for conversational machine comprehension. In: Proceedings of the Conference
  on Empirical Methods in Natural Language Processing, {EMNLP} 2019, pp 86--90,
  \doi{10.18653/v1/D19-5812}

\bibitem[{Zaib et~al.(2020)Zaib, Sheng, and
  Zhang}]{45663f8dbad442a7b649001bfeb2be72}
Zaib M, Sheng QZ, Zhang WE (2020) {A short survey of pre-trained language
  models for conversational AI: A new age in NLP}. In: Proceedings of the
  Australasian Computer Science Week Multiconference 2020, ACSW 2020, pp 1--4,
  \doi{10.1145/3373017.3373028}

\bibitem[{Zaib et~al.(2021)Zaib, Tran, Sagar, Mahmood, Zhang, and
  Sheng}]{10.1007/978-981-16-0010-4_5}
Zaib M, Tran DH, Sagar S, Mahmood A, Zhang WE, Sheng QZ (2021) {BERT-CoQAC:
  BERT-Based Conversational Question Answering in Context}. In: Parallel
  Architectures, Algorithms and Programming, pp 47--57

\bibitem[{Zellers et~al.(2018)Zellers, Bisk, Schwartz, and
  Choi}]{DBLP:conf/emnlp/ZellersBSC18}
Zellers R, Bisk Y, Schwartz R, Choi Y (2018) {SWAG:} a large-scale adversarial
  dataset for grounded commonsense inference. In: Proceedings of the Conference
  on Empirical Methods in Natural Language Processing, {EMNLP} 2018, pp
  93--104, \doi{10.18653/v1/d18-1009}

\bibitem[{Zhang et~al.(2018)Zhang, Chen, Ai, Yang, and
  Croft}]{zhang2018towards}
Zhang Y, Chen X, Ai Q, Yang L, Croft WB (2018) Towards conversational search
  and recommendation: system ask, user respond. In: Proceedings of the 27th acm
  international conference on information and knowledge management, {CIKM}
  2018, pp 177--186, \doi{10.1609/aaai.v34i03.5644}

\bibitem[{Zhong et~al.(2019)Zhong, Tang, Duan, Zhou, Wang, and
  Yin}]{DBLP:conf/nlpcc/ZhongTDZWY19}
Zhong W, Tang D, Duan N, Zhou M, Wang J, Yin J (2019) Improving question
  answering by commonsense-based pre-training. In: Proceedings of the 8th
  International Natural Language Processing and Chinese Computing Conference,
  {NLPCC} 2019, pp 16--28, \doi{10.1007/978-3-030-32233-5\_2}

\bibitem[{Zhu et~al.(2018)Zhu, Zeng, and
  Huang}]{DBLP:journals/corr/abs-1812-03593}
Zhu C, Zeng M, Huang X (2018) {SDNet}: contextualized attention-based deep
  network for conversational question answering. arXiv:181203593

\bibitem[{Zhu et~al.(2019)Zhu, Zeng, and Huang}]{zhu2019multi}
Zhu C, Zeng M, Huang X (2019) Multi-task learning for natural language
  generation in task-oriented dialogue. In: Proceedings of the Conference on
  Empirical Methods in Natural Language Processing and the 9th International
  Joint Conference on Natural Language Processing, {EMNLP-IJCNLP} 2019, pp
  1261--1266

\bibitem[{Zou(2020)}]{Zou_2020}
Zou X (2020) A survey on application of knowledge graph. Journal of Physics:
  Conference Series 1487:012016, \doi{10.1088/1742-6596/1487/1/012016}

\end{thebibliography}

%
%

\end{document}